\documentclass{article}

\usepackage{todonotes}
\usepackage{geometry}
\usepackage{amsmath}
\usepackage{amssymb}
\usepackage{amsthm}
\usepackage{bm}
\usepackage{graphicx}
\graphicspath{{./figures/}}
\usepackage{cite}
\usepackage{color}
\usepackage{tabu}
\usepackage{booktabs}
\usepackage{caption}
\usepackage{authblk}
\usepackage{subfigure}
\usepackage{enumerate}
\usepackage{tabu}
\usepackage{multirow}
\usepackage{diagbox}

\newcommand{\R}{\mathbb{R}}
\newcommand{\LO}{\mathcal{L}}
\newcommand{\LN}{\text{LayerNorm}}
\newcommand{\eps}{\varepsilon}

\newcommand*\diff{\mathop{}\!\mathrm{d}}

\begin{document}
\title{Capturing the Diffusive Behavior of the Multiscale Linear Transport Equations by Asymptotic-Preserving Convolutional DeepONets }

\author[1]{Keke Wu}
\author[1,3]{Xiong-bin Yan}
\author[1,2,4]{Shi Jin}
\author[1,2,3, \thanks{Corresponding author: zhengma@sjtu.edu.cn}]{Zheng Ma}

\affil[1]{School of Mathematical Sciences, Shanghai Jiao Tong University, Shanghai,
    200240, P. R. China.}
\affil[2]{ Institute of Natural Sciences, MOE-LSC,
    Shanghai Jiao Tong University, Shanghai, 200240, P. R. China.}
\affil[3]{Qing Yuan Research Institute,
    Shanghai Jiao Tong University, Shanghai, 200240, China.}
\affil[4]{Shanghai Artificial Intelligence Lab, Shanghai, China }

\date{\today}

\maketitle
\tableofcontents

\begin{abstract}

In this paper, we introduce two types of novel Asymptotic-Preserving Convolutional Deep Operator Networks (APCONs) designed to solve the multiscale time-dependent linear transport equations. We observe that the vanilla physics-informed DeepONets with modified MLP may exhibit instability in maintaining the desired limiting macroscopic behavior. Therefore, this necessitates the utilization of an asymptotic-preserving loss function. Drawing inspiration from the heat kernel in the diffusion equation, we propose a new architecture called Convolutional Deep Operator Networks, which employs multiple local convolution operations instead of a global heat kernel, along with pooling and activation operations in each filter layer. Our APCON methods possess a parameter count that is independent of the grid size and are capable of capturing the diffusive behavior of the linear transport problem. Finally, we validate the effectiveness of our methods through several numerical examples.

\end{abstract}

\section{Introduction}
% \todo{Review: (1) Multiscale kinetic problems; (2) DNN to solve PDEs, including PINN, DeepRitz, Operator learning, etc.}

 Kinetic equations model  the dynamics of particles that transport and  collide between themselves or interact with the background media or external fields.
These equations are defined in the phase space thus suffer from curse-of-dimensionality. In addition, they often possess multiple spatial and/or temporal scales, as well as non-local operators, posing notable hurdles for numerical simulations. 
{For a comprehensive exploration of this topic, see ~\cite{BGP,DP-Acta,jin2010asymptotic,weinan2011principles}.}

Deep learning methods and deep neural networks (DNNs) have received considerable attention within the scientific community, primarily due to their remarkable ability in addressing the intricate complexities involved in solving partial differential equations (PDEs)~\cite{beck2020, cai2021least, E2018, liao2019deep, lyu2020mim, raissi2019physics, deepGalerkin2018, zang2020weak, XDE}.
For those interested in exploring alternative machine learning approaches to tackle partial differential equations, we mention some exemplary review articles~\cite{beck2020, XDE}. 
These approaches are primarily motivated by the idea of parameterizing solutions or gradients of PDE problems using deep neural networks. 
As a result, they lead to a high-dimensional and non-convex minimization problem. Unlike traditional numerical methods, deep learning methods are mesh-free, enabling the solution of PDEs in intricate domains and geometries, and alleviating the issue of curse-of-dimensionality. 
The advantages lie in their flexibility and ease of implementation.
However, it is important to note that deep learning methods come with certain potential drawbacks, such as lengthy training times, convergence issues, and reduced accuracy.
While the majority of preceding neural network architectures excel in acquiring mappings between finite-dimensional Euclidean spaces, their capacity for generalization becomes constrained when confronted with disparate parameters, initial conditions, or boundary conditions.
The concept of operator learning presents a promising approach to solving a specific class of PDEs by training the neural network only {\it once},  see for example \cite{li2020fourier, lu2021learning, zhang2021mod, li2021physics, wang2021learning, xiong2023koopman, zhang2022multiauto, liu2022ht, cao2023lno,xu2023transfer} and references therein. 
As an example, the Fourier neural operator (FNO) framework~\cite{li2020fourier} exhibits the remarkable capability of proficiently acquiring the mapping between infinite-dimensional spaces through input-output pairs.
However, it is crucial to acknowledge that there are still several unresolved aspects pertaining to the convergence theory.

% % \todo{Review: DNNs for kinetic equation.}

In recent years, there has been a wealth of research dedicated to the utilization of deep neural networks in addressing multiscale kinetic equations and hyperbolic systems. This body of work encompasses a range of studies, some of which are referenced in the citations~\cite{lou2021physics, CLM, HJJL, wuAPNN, lu2022solving, li2022model, bertaglia2022asymptotic1, bertaglia2022asymptotic2,xu2023transfer,li2023solving}. The importance of tackling kinetic problems that exhibit characteristics across multiple scales has been widely recognized.
As we are aware, there exist numerous options to construct the loss when presented with a partial differential equation (PDE). {For example, the variational formulation (DRM)~\cite{E2018}, the least-squares formulation (PINN, DGM)~\cite{raissi2019physics,deepGalerkin2018}, and the weak formulation (WAN)~\cite{zang2020weak}, among others.} 
Due to the presence of small scales, the conventional Physics-Informed Neural Networks (PINNs) can perform poorly when dealing with multiscale kinetic equations~\cite{lu2022solving, wuAPNN}. 
Consequently, an alternative approach for solving multiscale kinetic equations utilizing Deep Neural Networks involves formulating a loss function that can effectively capture the macroscopic behavior at the limit --referred to as the Asymptotic-Preserving (AP) loss. A computational method for multiscale kinetic equations is called AP if it mimics the asymptotic transition from the kinetic equation to its macrocropic diffusive or hydrodynamic limit \cite{jin2010asymptotic}, which has been a standard criterion for developing multiscale kinetic solvers that can capture efficiently the macrosopic behavior even if the small physical scale--the Knudsen number--is not numerically resolved. 
This, in turn, justifies the necessity of employing Asymptotic-Preserving Neural Networks (APNNs)~\cite{wuAPNN}.
In our previous works~\cite{wuAPNN, wuAPNNv2}, we presented two distinct methodologies employing APNNs for addressing time-dependent linear transport equations with diffusive scaling and uncertainties. These approaches are based on micro-macro and even-odd decomposition techniques, respectively. Furthermore, we provided empirical evidence demonstrating the AP characteristics of the loss function as the Knudsen number tends towards zero.

% \todo[inline]{Our contributions}

This paper focuses on {\it operator learning} for the multiscale time-dependent linear transport equation. {The {\it operator learning} tasks include methods that endeavor to map infinite-dimensional function spaces by acquiring knowledge of solution operators through neural networks. These operators facilitate the mapping of input functions (e.g., initial conditions, boundary conditions, or forcing terms) to solutions of partial differential equations~\cite{lu2021learning, li2020fourier}.}
It has been observed that the vanilla physics-informed DeepONets, when equipped with a modified {multi-layer perceptron}, may encounter instability issues in preserving the desired macroscopic behavior. 
To address this, we first introduce the concept of an AP loss function in the context of operator learning. Subsequently, taking inspiration from the heat kernel in the diffusion equation, we propose a novel architecture named Convolutional DeepONets. 
These networks employ multiple local convolution operations, instead of a global heat kernel, along with pooling and activation operations in each filter layer. 
As a result, we introduce two innovative types of Asymptotic-Preserving Convolutional DeepONets (APCONs) based on micro-macro and even-odd decomposition--which are two popular AP schemes for multiscale linear transport equations with diffusive scaling-- respectively. 
Notably, our APCON methods possess a parameter count that remains unaffected by the grid size, 
while effectively capturing the diffusive behavior inherent in the linear transport problem.

% % \todo[inline]{Structure of the  paper.}
An outline of this paper is as follows.
In Section 2, a detail illustration of the linear transport equations under the diffusion regime are given.
Section 3 is our main part, we reviewed the physics-informed DeepONets and proposed two novel Asymptotic-Preserving Convolutional DeepONets for the multiscale time-dependent linear transport problem.
Numerous numerical examples are presented in Section 4 to demonstrate the effectiveness of the APCONs.
The paper is concluded in Section 5.

\section{The linear transport equations under the diffusion regime}

We investigate the linear transport problem within the framework of diffusive scaling, specifically in the context of one-dimensional space and velocity variables. The problem is formulated as follows:
\begin{equation}\label{eqn: linear-transport-1d}
    \eps \partial_t f + v \cdot \nabla_x f = \frac{1}{\eps}\LO f + \eps Q,
\end{equation}
with corresponding initial and boundary conditions.

In this context, the function $f(t, x, v)$ represents the density distribution of particles at a given time $t \in \mathcal{T} := [0, T]$, at a spatial location $x \in \mathcal{D} := [x_L, x_R]$, and moving in the direction $v \in \Omega := [-1, 1]$. 
The parameter $\eps > 0$ corresponds to the Knudsen number,  the ratio of the mean free path to a characteristic length. 
Furthermore, $Q(t, x)$ denotes the source function, $\LO$ is the linear collision operator given by 
\begin{equation}\label{eqn: collison-operator}
    \LO f = \frac{1}{2} \int_{-1}^1 f \, \diff v - f.
\end{equation}

\subsection{The micro-macro decomposition}

In this section, we present the micro-macro decomposition formulation for the linear transport equation~\cite{lemou2008new}. 
The approach involves decomposing the function $f$ into two components: the equilibrium part denoted as $\rho(t, x)$ and the non-equilibrium part represented by $g(t, x, v)$:
\begin{equation}\label{eqn: micro-macro-decomposition}
    f = \rho + \eps g,
\end{equation}
where
\begin{equation}\label{eqn: constraint-rho}
    \rho := \left \langle f \right \rangle = \frac{1}{2}\int_{-1}^{1} f \, \diff{v}.
\end{equation}
The non-equilibrium component $g$ satisfies the following relation:
\begin{equation}\label{eqn: constraint-g}
    \left \langle g \right \rangle = 0.
\end{equation}
By applying equation \eqref{eqn: micro-macro-decomposition} to equation \eqref{eqn: linear-transport-1d}, one obtains the following expression:
\begin{equation}\label{eqn: linear-transport-decomposition}
    \eps \partial_t \rho + \eps^2  \partial_t g + v \cdot  \nabla_{x} \rho + \eps v \cdot \nabla_{x} g = \LO g + \eps Q.
\end{equation}
By integrating this equation with respect to $v$,  the following continuity equation with a source term can be derived:
\begin{equation}\label{eqn: macro}
    \partial_t \rho + \nabla_{{x}} \cdot \left \langle  {v}g \right \rangle = Q.
\end{equation}
By defining the operator $\Pi$ as $\Pi(\cdot)({v}) = \left\langle \cdot \right\rangle$ and $I$ as the identity operator, one  obtains an evolution equation for $g$ by applying the orthogonal projection $I - \Pi$ to equation \eqref{eqn: linear-transport-decomposition}.
\begin{equation}\label{eqn: micro}
    \eps^2 \partial_t g + \eps (I - \Pi)({v} \cdot \nabla_{{x}} g) + {v} \cdot  \nabla_{{x}} \rho = \mathcal{L} g + (I - \Pi) \eps Q.
\end{equation}
The micro-macro formulation of equation \eqref{eqn: linear-transport-1d} for our APCON consists of equations \eqref{eqn: macro} and \eqref{eqn: micro}, along with the constraints \eqref{eqn: constraint-rho} and \eqref{eqn: constraint-g}:
\begin{equation}\label{eqn: micro-macro}
    \left\{
    \begin{aligned}
         & \partial_t \rho + \nabla_{{x}} \cdot \left \langle  {{v}}g \right \rangle = Q,                                                                               \\
         & \eps^2 \partial_t g + \eps (I - \Pi)({v} \cdot \nabla_{{x}} g) + {v} \cdot  \nabla_{{x}} \rho = \LO g + (I - \Pi) \eps Q, \\
         & \langle g \rangle = 0.
    \end{aligned}
    \right.
\end{equation}
As $\eps$ tends to 0, the aforementioned system formally approaches the following:
\begin{equation}\label{eqn: ap-limit}
    \left\{
    \begin{array}{ll}
        \partial_t \rho + \nabla_{{x}} \cdot \left \langle  {{v}}g \right \rangle = Q, \\
        {v} \cdot  \nabla_{{x}} \rho = \mathcal{L} g .
    \end{array}
    \right.
\end{equation}
The second equation yields
\begin{equation}\label{eqn: inverse-operator}
    g=\mathcal{L}^{-1}({v} \cdot  \nabla_{{x}} \rho),
\end{equation}
upon substituting the expressions into the first equation and integrating over $v$, one obtains the diffusion equation \eqref{eqn: diffusion-limit}:
\begin{equation}\label{eqn: diffusion-limit}
    \partial_t \rho = \frac{1}{3} \partial_{xx} \rho  + Q\,.
\end{equation}

\subsection{The even-odd decomposition}

An alternative approach, known as the even-odd decomposition, was introduced by Lewis and Miller~\cite{lewis1984computational}. This method involves splitting equation \eqref{eqn: linear-transport-1d} into two separate equations, one for $v$ and another for $-v$:
\begin{subequations}\label{eqn: parity}
\begin{equation}
    \eps \partial_t f(v) + v \cdot \nabla_x f(v) = \frac{1}{\eps}\LO f(v) + \eps Q,
\end{equation}

\begin{equation}
    \eps \partial_t f(-v) - v \cdot \nabla_x f(-v) = \frac{1}{\eps}\LO f(-v) + \eps Q,
\end{equation}
\end{subequations}
and let us define even- and odd-parities as follows:
\begin{equation}
    \begin{aligned}
        r(t, x, v) & = \frac{1}{2}[f(t, x, v) + f(t, x, -v)],             \\
        j(t, x, v) & = \frac{1}{2\eps}[f(t, x, v) - f(t, x, -v)].
    \end{aligned}
\end{equation}
By adding and subtracting the two equations in equation \eqref{eqn: parity}, one can derive the following system of equations:
\begin{equation}\label{eqn: even-odd}
    \begin{aligned}
         & \partial_t r + v \cdot \nabla_x j = \frac{1}{\eps^2} \LO r + Q,                   \\
         & \partial_t j + \frac{1}{\eps^2} v \cdot \nabla_x r = -\frac{1}{\eps^2} j.
    \end{aligned}
\end{equation}
The first equation by integrating over $v$ gives
\begin{equation}\label{eqn: integral-equation-1}
    \partial_t \left \langle r \right \rangle  + \left \langle v \cdot \nabla_x j \right \rangle = \frac{1}{\eps^2} (\rho - \left \langle r \right \rangle) + Q,
\end{equation}
and since $\rho = \left\langle r \right\rangle$, one can express it as follows:
\begin{equation}\label{eqn: integral-equation-2}
    \partial_t \rho + \left \langle v \cdot \nabla_x j \right \rangle = Q.
\end{equation}
The even-odd formulation of equation \eqref{eqn: linear-transport-1d} for our APCON consists of equations (\ref{eqn: even-odd}) and (\ref{eqn: integral-equation-2}), along with the constraint $\rho = \left \langle r \right \rangle$:
\begin{equation}\label{eqn: even-odd-system}
    \left \{
    \begin{aligned}
         & \eps^2 \partial_t r + \eps^2 v \cdot \nabla_x j = \rho - r + \eps^2 Q, \\
         & \eps^2 \partial_t j + v \cdot \nabla_x r = - j,                   \\
         & \partial_t \rho +  \left \langle v \cdot \nabla_x j \right \rangle = Q,  \\
         & \rho = \left \langle r \right \rangle.
    \end{aligned}
    \right.
\end{equation}
When $\eps \to 0$, the above equation formally approaches
\begin{equation}\label{eqn: AP-limit}
    \left \{
    \begin{aligned}
         & r = \rho,                                                           \\
         & j = - v \cdot \nabla_x r,                                               \\
         & \partial_t \rho +  \left \langle v \cdot \nabla_x j \right \rangle = Q.
    \end{aligned}
    \right.
\end{equation}
By substituting the first equation into the second equation, we obtain $j = - v \partial_x \rho$. Plugging this expression into the third equation leads to the diffusion equation \eqref{eqn: diffusion-limit}.

\subsection{The heat kernel of diffusion equation}

A fundamental solution, commonly referred to as a heat kernel, is a solution of the diffusion or heat equation that corresponds to an initial condition where heat is concentrated at a specific point. 
These fundamental solutions are valuable in determining general solutions of the heat equation across specific domains. 
For further exploration of this topic, introductory treatments can be found in references such as \cite{strauss2007partial, evans2010}.

Taking the one-dimensional case as an example, the Green's function serves as a solution to the initial value problem. 
By utilizing Duhamel's principle, this is equivalent to defining the Green's function as the solution to the first equation, where a delta function is employed.

\begin{equation}\label{eqn: heat}
    \left \{
    \begin{aligned}
         & u_t - k u_{xx} = 0,     \; (t, x) \in \R^+ \times \R        \\
         & u(0,x) = \delta(x),
    \end{aligned}
    \right.
\end{equation}
where $\delta(x)$ represents the Dirac delta function. 
The solution obtained through the Fourier transform corresponds to the fundamental solution, also known as the heat kernel of equation \eqref{eqn: heat}:
\begin{equation}\label{eqn: fundamental-solution}
    \Phi(t, x) = \frac{1}{\sqrt{4 \pi k t}} \exp \left ( - \frac{x^2}{4 k t} \right ).
\end{equation}
The general solution of the one-variable heat equation with an initial condition $u(0,x) = g(x)$, where $(t, x) \in \mathbb{R}^+ \times \mathbb{R}$, can be obtained by employing a convolution operation, known as the Poisson formula:
\begin{equation}\label{eqn: convolution}
    u(t, x) = \int_{-\infty}^{\infty} \Phi(t, x - y) g(y) \diff{y}.
\end{equation}
In the case of multiple spatial variables, the fundamental solution solves the analogous problem, and the fundamental solution in $n$ variables is obtained by taking the product of the fundamental solutions in each individual variable:
\begin{equation}\label{eqn: fundamental-solution-nd}
    \Phi(t, \bm{x}) = \frac{1}{{(4 \pi k t)}^{\frac{n}{2}}} \exp \left ( - \frac{|\bm{x}|^2}{4 k t} \right ).
\end{equation}
The general solution of the heat equation on $\mathbb{R}^d$ can be obtained through convolution, allowing one to solve the initial value problem with the condition $u(0, \bm{x}) = g(\bm{x})$, where $\bm{x}$ represents a vector in $\mathbb{R}^d$.

Consider the inhomogeneous case on the whole line as 
\begin{equation}\label{eqn: heat-inhomogeneous}
    \left \{
    \begin{aligned}
         & u_t - k u_{xx} = f(t, x),     \; (t, x) \in \R^+ \times \R        \\
         & u(0,x) = g(x),
    \end{aligned}
    \right.
\end{equation}
with $f(t, x), g(x)$ arbitrary given functions. 
For example, if $u(t, x)$ represents the temperature distribution along a rod, then $g(x)$ corresponds to the initial temperature distribution along the rod, and $f(t, x)$ represents a heat source (or sink) applied to the rod at later times.

Using Duhamel's principle, we can obtain the solution of equation \eqref{eqn: heat-inhomogeneous} as follows:
\begin{equation}\label{eqn: convolution-inhomogeneous}
    u(t, x) = \int_{-\infty}^{\infty} \Phi(t, x - y) g(y) \diff{y} + \int_0^t \int_{-\infty}^{\infty} \Phi(t - s, x - y) f(s, y) \diff{y} \diff{s}.
\end{equation}

\section{Operator learning for the linear transport equation}

Let us consider the multiscale time-dependent linear transport equation with diffusive scaling, which can be expressed as follows:
\begin{equation}\label{eqn: lte}
    \left \{
        \begin{aligned}
         & \eps \partial_t f + v \cdot \nabla_x f = \frac{1}{\eps}\LO f, \; (t, x, v) \in  \mathcal{T} \times \mathcal{D} \times {\Omega}, \\
         & \mathcal{B} f(t, x, v) = 0,      \\
         & f(0, x, v) = f_0(x, v).
    \end{aligned}
    \right.
\end{equation}
Our objective is to train a solution neural operator that maps the initial function $f_0(x, v)$ to the corresponding solutions of the PDE, denoted as $f(t, x, v)$.

It is important to note that the function space of $f_0(x, v)$ is infinite-dimensional. To represent it using neural networks, we need to utilize finite function values at specific points in order to construct a finite-dimensional approximation.

\subsection{Neural networks}

First, we introduce the notations for conventional multi-layer percerptron (MLP) networks\footnote{BAAI.2020.\ Suggested Notation
    for Machine Learning.\ https://github.com/mazhengcn/suggested-notation-for-machine-learning.}.
An $L$-layer MLP is defined recursively as,
\begin{equation}
    \begin{aligned}
        f^{[0]}(x) & = x,                                                                              \\
        f^{[l]}(x) & = \sigma \circ (W^{[l-1]} f^{[l-1]}(x) + b^{[l-1]}), \, 1 \le l \le L-1, \\
        f_{\theta}(x)       & = W^{[L-1]} f_{\theta}^{[L-1]}(x) + b^{[L-1]},
    \end{aligned}
\end{equation}
% here, $W^{[l]} \in  \R^{m_{l+1}\times m_l}, b^{l}\in  \R^{m_{l+1}}, m_0, m_{L}$ are the networks input and output dimension and $\sigma$ is a nonlinear scalar function. The notation ``$\circ$'' means entry-wise operation. We use the same notations in the following context for convenience if it does not cause any confusion. 
here, we denote $W^{[l]} \in \mathbb{R}^{m_{l+1} \times m_l}$ and $b^{[l]} \in \mathbb{R}^{m_{l+1}}$ as the weight matrix and bias vector, respectively, for the $l$-th layer of the MLP. The dimensions $m_0$ and $m_L$ correspond to the input and output dimensions of the network. The symbol ``$\sigma$" represents a scalar nonlinear activation function. The notation ``$\circ$" denotes an element-wise operation. We will use these notations consistently throughout the subsequent discussions for convenience, as long as it does not lead to any ambiguity.

In this paper, we employ the modified MLP architecture with layer normalization~\cite{wang2021understanding, ba2016layer}. 
The architecture can be expressed as follows:
\begin{align}
    U &= \sigma \circ(W_1^{[1]} x + b_1^{[1]}), V = \sigma \circ(W_2^{[1]} x + b_2^{[1]}), \\
    H^{[1]} & = \sigma \circ(W^{[1]} x + b^{[1]}), \\
    Z^{[l]} & = \sigma \circ(\LN(W^{[l]} H^{[l]} + b^{[l]})), \ \ l = 1, \dots, L-1, \\
    H^{[l+1])} & = (1 - Z^{[l]}) \odot U  +  Z^{[l]}  \odot V, \ \  l = 1, \dots, L-1, \\
    f_{\theta}(x) & = W^{[L]} H^{[L]} + b^{[L]},
\end{align}
where $W_1^{[1]}, W_2^{[1]} \in \R^{m_{1}\times m_0},  b_1^{[1]}, b_2^{[1]} \in \R^{m_{1}}$ and $\odot$ denotes element-wise multiplication. {In the study by Wang et al.~\cite{wang2021understanding}, the incorporation of residual connections and consideration of multiplicative interactions among the inputs were shown to result in enhanced predictive performance. Layer normalization indeed constitutes a valuable technique employed to normalize the distributions of intermediate layers. Through this approach, it facilitates the acquisition of smoother gradients, thus expediting the training process and culminating in heightened generalization accuracy.}

\subsection{Physics-informed DeepONets}

The concept of learning operators through a parametric-based approach was introduced in \cite{chen1995universal}, where the authors proposed a method utilizing a one-layer MLP to learn non-linear operators. Furthermore, they presented a universal approximation theorem, guaranteeing that their architecture can approximate any continuous operator with arbitrary precision. Building upon this work, DeepONet was introduced in \cite{lu2021learning}, employing a multi-layer MLP and demonstrating its efficacy in approximating solution operators for a wide range of ordinary and partial differential equations.

\begin{figure}[ht]
    \centering
    \includegraphics[width=0.75\textwidth]{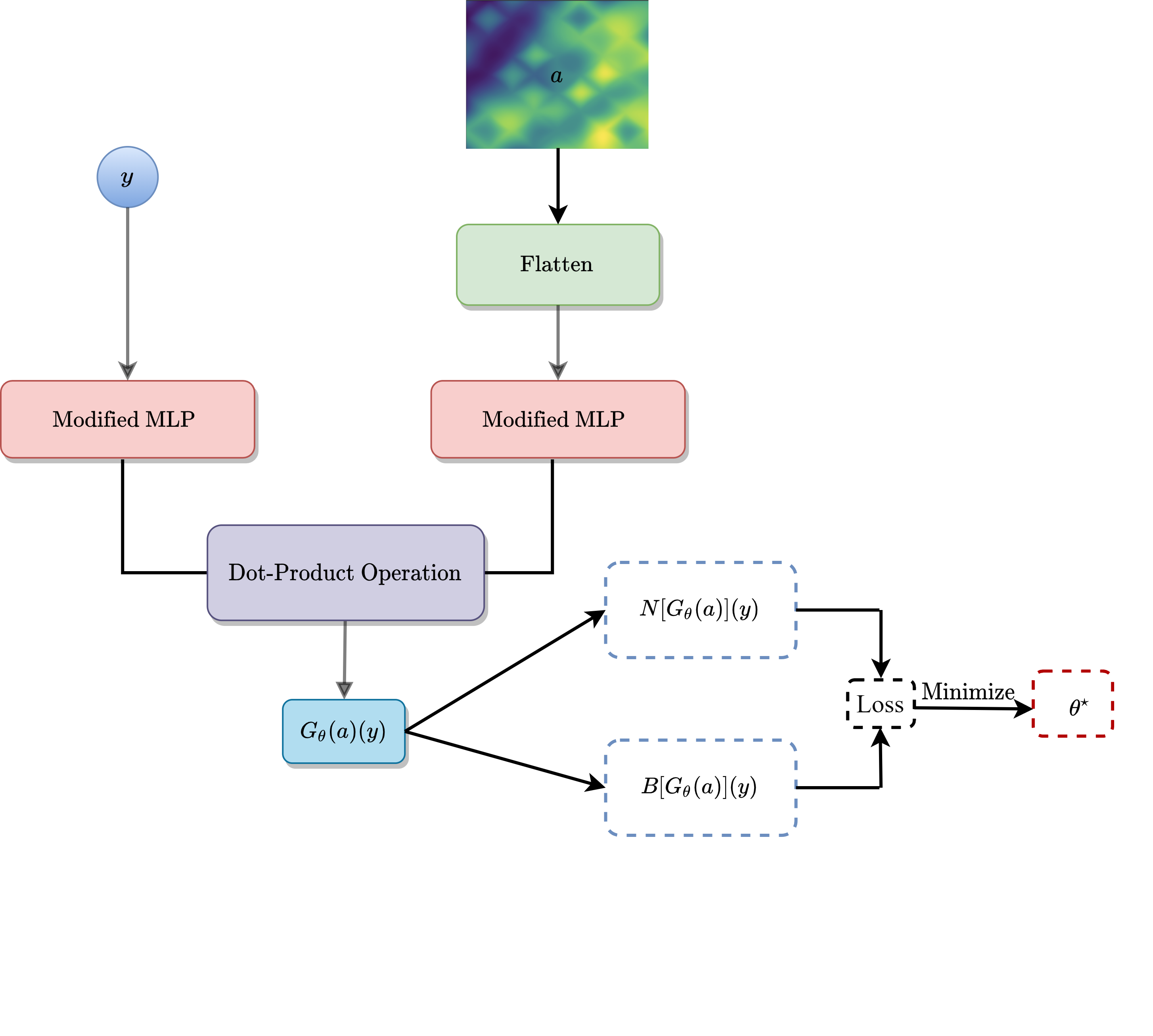}
    \caption{{The architecture of the Physics-informed DeepONets} is characterized by its core component, the DeepONet with a modified MLP. This architecture includes two distinct sub-networks dedicated to extracting underlying representations of input functions denoted as $a$, and the corresponding coordinates denoted as $y$. Moreover, in relation to the equation component, the loss function for the PDE is denoted as $N[G_\theta(a)](y)$, while the loss function for the boundary, initial, and other observed data is denoted as $B[G_\theta(a)](y)$.}
    \label{fig: general-pidons}
\end{figure}

A DeepONet consists of two sub-networks: the branch net, responsible for encoding the input function $a(x)$ at a fixed number of sensors $x_i \in \mathbb{R}^d$, where $i = 1, \cdots, m$; and the trunk net, responsible for encoding the locations $y$ of the output functions. The objective of DeepONet is to learn an operator $G: a(y) \mapsto G(a)(y)$, which takes two inputs: $[a(x_1), a(x_2), \cdots, a(x_m)]$ and $y \in \mathbb{R}^d$.
Assuming that the branch net and trunk net output two vectors of length $p$: $[b_1, b_2, \cdots, b_p]^T \in \mathbb{R}^p$ and $[t_1, t_2, \cdots, t_p]^T \in \mathbb{R}^p$, respectively, the approximate output $G(a)(y)$ can be obtained as $\sum_{j=1}^p b_j t_j$. Additionally, a bias term $b_0 \in \mathbb{R}$ can be incorporated in the final stage if desired:
\begin{equation}
    G(a)(y) \approx \sum_{j=1}^p b_j t_j + b_0.
\end{equation}
In \cite{wang2021learning}, an extension of DeepONet called physics-informed DeepONets (PIDONs) was introduced. PIDONs incorporate a regularization term into the loss function to enforce known physical constraints. This regularization enables the prediction of solutions to parametric differential equations, even in cases where paired input-output training data is not available.
We have included a comprehensive schematic diagram illustrating the PIDON methodology employed for resolving the PDE in Figure~\ref{fig: general-pidons}.

Physics-informed DeepONets leverage available data and/or physical constraints to acquire knowledge regarding the solution operator for a particular set of PDEs. By doing so, they are capable of surmounting the limitations associated with purely data-driven approaches and those solely based on physics principles. The neural network architecture employed in physics-informed DeepONets, as previously mentioned, involves a modified MLP~\cite{wang2021understanding} which has been specifically tailored for this purpose.

In the context of solving the linear transport equation using neural operators, the physics-informed DeepONet methodology employs the flattened and discretized initial function denoted as $a$, along with the location coordinates $(t, x, v)$, as inputs for both the branch and trunk networks. These networks then generate the approximate neural solution denoted as $f_\theta$, which is evaluated at the corresponding coordinates $(t, x, v)$.

Consequently, a single DeepONet, equipped with a modified MLP, is utilized to construct the solution operator for a specific input denoted as $a:$
\begin{equation}\label{eqn: pidon-solution}
    f_\theta[a](t, x, v) := \sigma_+ (G_\theta(a)(t, x, v)) \approx f(t, x, v),
\end{equation}
here $\sigma_+(\cdot) = \log (1 + \exp(\cdot))$ is applied to keep $f_\theta[a](t, x, v)$ be positive.

The loss function of the linear transport equation for the Physics-informed DeepONet, considering a specific input denoted as $a$, can be expressed as follows:
\begin{equation}\label{eqn: single-pidon-loss}
    \begin{aligned}
    \mathcal{L}_{\text{\tiny{PIDON}}}^a & = \frac{1}{|\mathcal{T} \times \mathcal{D} \times \Omega|} \int_{\mathcal{T} \times \mathcal{D} \times \Omega} |\eps^2 \partial_t f_\theta + \eps v \cdot \nabla_x f_\theta - \LO f_\theta |^2 \diff{t} \diff{x} \diff{v}  \\
                               & + \frac{1}{| \mathcal{T} \times \Omega|} \int_{\mathcal{T} \times \Omega} | \mathcal{B} f_\theta[a](t, x, v) - 0 |^2 \diff{t} \diff{v} \\
                               & + \frac{1}{| \mathcal{D} \times \Omega|} \int_{\mathcal{D} \times \Omega} | f_\theta[a](0, x, v) - f_0(x, v) |^2 \diff{x} \diff{v},                 
    \end{aligned}
\end{equation}
where $| X |$ denotes the measure of the domain $X$. 
In the operator learning task, it is necessary to compute the average of $\mathcal{L}_{\small{\text{PIDON}}}^a$ across all samples $a \in \{ a^i\}_{i=1}^M$. The corresponding total loss function is defined as follows:
\begin{equation}\label{eqn: pidon-loss}
   \mathcal{L}_{\text{\tiny{PIDON}}} = \frac{1}{M} \sum_{i=1}^M \mathcal{L}_{\text{\tiny{PIDON}}}^{a^i}.
\end{equation}

\begin{figure}[ht]
    \centering
    \includegraphics[width=0.75\textwidth]{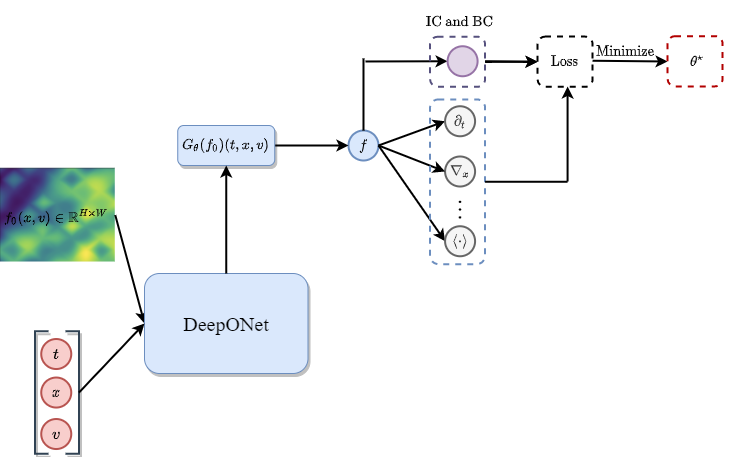}
    \caption{The architecture of Physics-informed DeepONets for linear transport equation.}
    \label{fig: pidons}
\end{figure}

We put a schematic plot of PIDON method for the linear transport equation in Figure~\ref{fig: pidons}.

\subsection{Asymptotic-Preserving Convolutional DeepONets}

\subsubsection{Convolutional DeepONets}

Although DeepONets still rely on large annotated datasets comprising paired input-output observations, they offer a straightforward and intuitive model that enables a continuous representation of the target output functions, regardless of resolution. However, it is worth noting that the sensors of the branch network in DeepONets typically operate on grid points within the domain in $\mathbb{R}^d$. As the dimensionality increases, this approach leads to an exponential growth in the number of grid points and consequently leads to a substantial increase in the size of neural parameters within the first layer of the branch network.

\begin{figure}[ht]
    \centering
    \includegraphics[width=0.75\textwidth]{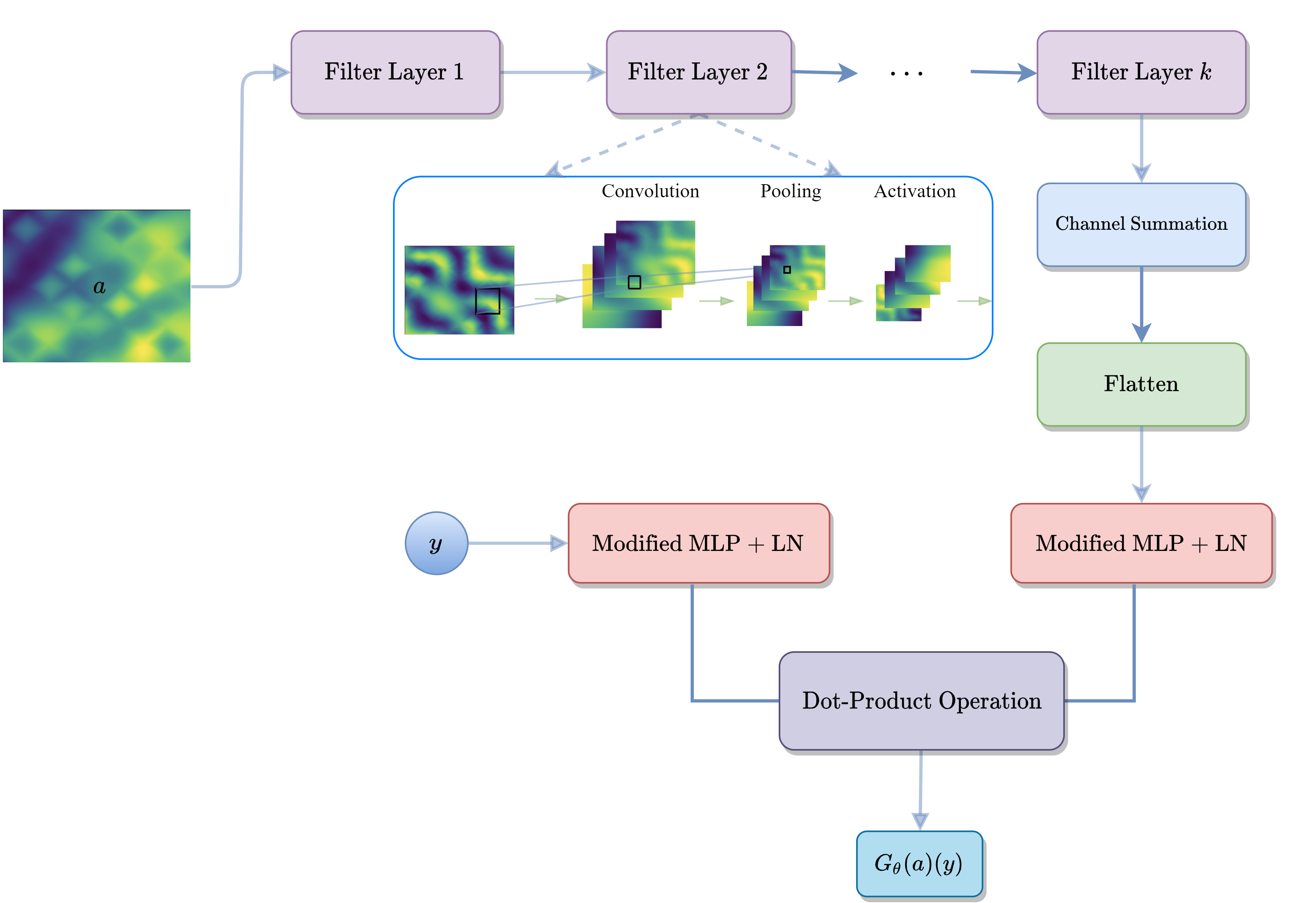}
    \caption{The architecture of the 2D Convolutional DeepONets: (a) Branch net - Commence with an input function $a$ having a discretized shape of $(H^{\text{in}}, W^{\text{in}})$. 1. Implement a series of filter layers, including convolution with $C$ channels, pooling, and activation operations, with the resulting shape being $(C, \; H^{\text{out}}, W^{\text{out}})$. The order of pooling and activation operations can be adjusted as desired. 2. Sum the outputs from the filter layers across the channels, resulting in a flattened output of shape $(1, \; H^{\text{out}} W^{\text{out}})$. 3. Elevate the flattened output to a higher-dimensional space using a one-layer MLP and feed it into the modified MLP with layer normalization. (b) Trunk net - Begin with a single point denoted as $y$. Apply the modified MLP with layer normalization to process the input.} 
    \label{fig: cons}
\end{figure}

Taking inspiration from the macroscopic limit of the linear transport equation as $\eps \to 0$ and the Poisson formula of the diffusion equation (Equations \eqref{eqn: convolution} and \eqref{eqn: convolution-inhomogeneous}), we propose a novel architecture called Convolutional DeepONets (CONs). This architecture utilizes a sequence of convolution, pooling, and activation operations (referred to as filter layers) for the inputs. Following these filter layers, we perform a summation along the channels and flatten the resulting output. Optionally, one can perform pooling and activation operations in a different order. Subsequently, the output is passed through the modified MLP with layer normalization. The underlying concept behind this approach is to employ multiple local convolution operations instead of a global heat kernel, thereby reducing the parameter size of the branch network.
In addition, we apply the modified MLP with layer normalization for the trunk network. Notably, the number of parameters in our Convolutional DeepONets is not dependent on the grid size. The architecture of the 2D Convolutional DeepONets is illustrated in Figure~\ref{fig: cons}. It is important to mention that this diagram depicts the specific structure for the 2D case, while a more general structure for higher-dimensional cases can be constructed following a similar analogy.

In the study conducted in \cite{zhang2022multiauto}, the authors introduced a model called MultiAuto-DeepONet. This model utilizes an encoder-decoder architecture to handle high-dimensional input random processes. The dimensionality reduction process in their model is achieved solely through convolutional layers acting as encoders, in contrast to our approach which incorporates both convolutional layers and a vanilla DeepONet with MLP. The difference arises from the intention to capture the diffusive characteristics of multiscale time-dependent linear transport equations in our model.

\subsubsection{APCONs based on the micro-macro decomposition}

{The APCON method based on micro-macro decomposition} utilizes the discretized initial function $a$ along with the location coordinates $(t, x, v)$ or $(t, x)$ as inputs for both the branch and trunk networks. The method then generates the approximate neural solutions $\rho_\theta$ evaluated at $(t, x)$, while $g_\theta$ is evaluated at $(t, x, v)$.

To construct the solution operator for $\rho$ and $g$ with respect to a specific input $a$, two Convolutional DeepONets are employed. These Convolutional DeepONets are utilized to model and approximate the solution operators for $\rho$ and $g$ respectively,
\begin{equation}\label{eqn: apcon-v1-solution}
    \begin{aligned}
        & \rho_\theta[a](t, x) := G_\theta(a)(t, x) \approx \rho(t, x), \\
        & g_\theta[a](t, x, v) := G_\theta(a)(t, x, v) - \left \langle G_\theta(a)(t, x, v) \right \rangle \approx g(t, x, v).
    \end{aligned}
\end{equation}
% the reason for not being positve

The loss function of the linear transport equation for APCON, based on micro-macro decomposition and considering a specific input $a$, can be expressed as follows:
{
\begin{equation}\label{eqn: single-apcon-v1-loss}
    \begin{aligned}
    \mathcal{L}_{\text{\tiny{APCON-v1}}}^a & = \frac{1}{|\mathcal{T} \times \mathcal{D} |} \int_{\mathcal{T} \times \mathcal{D}} | \partial_t \rho_\theta + \left \langle v \cdot \nabla_x  g_\theta \right \rangle |^2 \diff{t} \diff{x}  \\
                                & + \frac{1}{|\mathcal{T} \times \mathcal{D} \times \Omega|} \int_{\mathcal{T} \times \mathcal{D} \times \Omega} |\eps^2 \partial_t g_\theta + \eps (I - \Pi)({v} \cdot \nabla_{{x}} g_\theta) + {v} \cdot  \nabla_{{x}} \rho_\theta - \mathcal{L} g_\theta |^2 \diff{t} \diff{x} \diff{v}  \\
                               & + \frac{1}{| \mathcal{T} \times \Omega|} \int_{\mathcal{T} \times \Omega} | \mathcal{B} (\rho_\theta + \eps g_\theta)[a](t, x, v) - 0 |^2 \diff{t} \diff{v} \\
                               & + \frac{1}{| \mathcal{D} \times \Omega|} \int_{\mathcal{D} \times \Omega} | (\rho_\theta + \eps g_\theta)[a](0, x, v) - f_0(x, v) |^2 \diff{x} \diff{v}.    
    \end{aligned}
\end{equation}
}
In the aforementioned construction, the constraint \eqref{eqn: constraint-g} is automatically omitted, similar to the approach described in the work ~\cite{wuAPNN}.

In the operator learning task, we compute the average of $\mathcal{L}_{\small{\text{APCON-v1}}}^a$ across all samples $a \in \{ a^i\}_{i=1}^M$ to obtain the corresponding total loss function, which is defined as follows:
\begin{equation}\label{eqn: apcon-v1-loss}
   \mathcal{L}_{\text{\tiny{APCON-v1}}} = \frac{1}{M} \sum_{i=1}^M \mathcal{L}_{\text{\tiny{APCON-v1}}}^{a^i}.
\end{equation}

\begin{figure}[ht]
    \centering
    \includegraphics[width=0.75\textwidth]{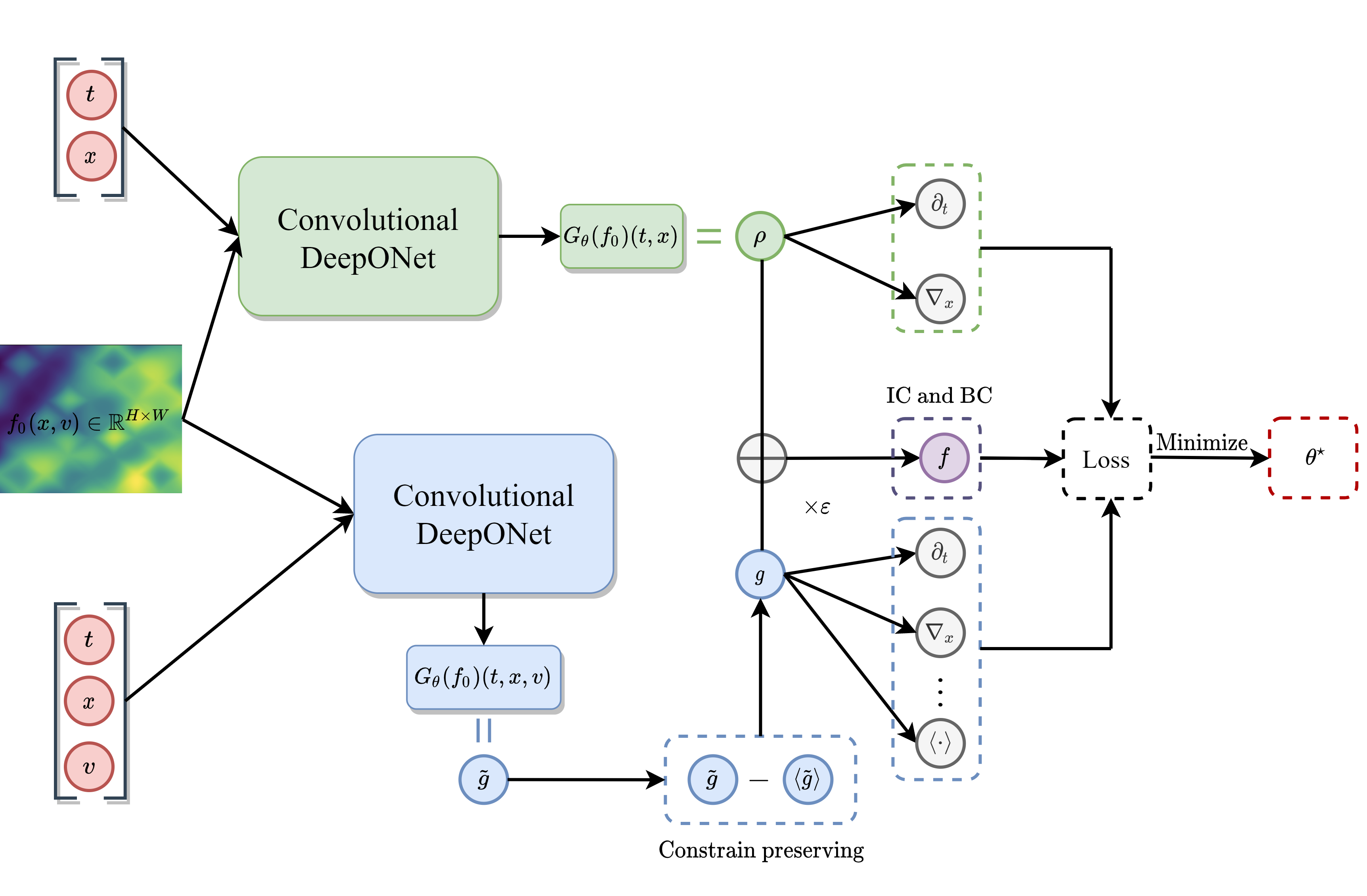}
    \caption{The architecture of Asymptotic-Preserving Convolutional DeepONets based on the micro-macro decomposition for linear transport equation.}
    \label{fig: apcons-micro-macro}
\end{figure}

We put a schematic plot of our APCON method based on micro-macro decomposition for the linear transport equation in Figure~\ref{fig: apcons-micro-macro}.

\subsubsection{APCONs based on the even-odd decomposition}

The APCON method, based on even-odd decomposition, utilizes the discretized initial function $a$ along with the location coordinates $(t, x, v)$ or $(t, x)$ as inputs for both the branch and trunk networks. The method then generates the approximate neural solutions $\rho_\theta$ evaluated at $(t, x)$, while $r_\theta$ and $j_\theta$ are evaluated at $(t, x, v)$.

To construct the solution operator for $\rho$, $r$ and $j$ with respect to a specific input $a$, three Convolutional DeepONets are employed. These Convolutional DeepONets are utilized to model and approximate the solution operators for $\rho$, $r$ and $j$ respectively,
\begin{equation}\label{eqn: apcon-v2-solution}
    \begin{aligned}
        & \rho_\theta[a](t, x) := G_\theta(a)(t, x) \approx \rho(t, x), \\
        & r_\theta[a](t, x, v) := \frac{1}{2} \left ( G_\theta(a)(t, x, v) + G_\theta(a)(t, x, -v) \right ) \approx r(t, x, v), \\
        & j_\theta[a](t, x, v) := G_\theta(a)(t, x, v) - G_\theta(a)(t, x, -v) \approx j(t, x, v).
    \end{aligned}
\end{equation}
Here $r_\theta[a](t, x, v)$ and $j_\theta[a](t, x, v)$ satisfy the even-odd property.

The loss function of the linear transport equation for APCON, based on the even-odd parity formulation and considering a specific input $a$, can be expressed as follows:
\begin{equation}\label{eqn: single-apcon-v2-loss}
    \begin{aligned}
    \mathcal{L}_{\text{\tiny{APCON-v2}}}^a & = \frac{1}{|\mathcal{T} \times \mathcal{D} \times \Omega |} \int_{\mathcal{T} \times \mathcal{D} \times \Omega } | \eps^2 \partial_t r_\theta + \eps^2 v \cdot \nabla_x  j_\theta - (\rho_\theta - r_\theta) |^2 \diff{t} \diff{x} \diff{v}  \\
                                & + \frac{1}{|\mathcal{T} \times \mathcal{D}  \times \Omega |} \int_{\mathcal{T} \times \mathcal{D} \times \Omega} | \eps^2 \partial_t j_\theta + v \cdot \nabla_x  r_\theta - ( - j_\theta) |^2 \diff{t} \diff{x} \diff{v}  \\
                                & + \frac{1}{|\mathcal{T} \times \mathcal{D} |} \int_{\mathcal{T} \times \mathcal{D}} | \partial_t \rho_\theta + \left \langle v \cdot \nabla_x j_\theta \right \rangle |^2 \diff{t} \diff{x}   \\
                                & + \frac{1}{|\mathcal{T} \times \mathcal{D} |} \int_{\mathcal{T} \times \mathcal{D}} | \rho_\theta - \left \langle r_\theta \right \rangle |^2 \diff{t} \diff{x}  \\
                               & + \frac{1}{| \mathcal{T} \times \Omega|} \int_{\mathcal{T} \times \Omega} | \mathcal{B} (r_\theta + \eps j_\theta)[a](t, x, v) - 0 |^2 \diff{t} \diff{v} \\
                               & + \frac{1}{| \mathcal{D} \times \Omega|} \int_{\mathcal{D} \times \Omega} | (r_\theta + \eps j_\theta)[a](0, x, v) - f_0(x, v) |^2 \diff{x} \diff{v}.    
    \end{aligned}
\end{equation}

In this operator learning task, we compute the average of $\mathcal{L}_{\small{\text{APCON-v2}}}^a$ across all samples $a \in \{ a^i\}_{i=1}^M$ to obtain the corresponding total loss function, which is defined as follows:
\begin{equation}\label{eqn: apcon-v2-loss}
   \mathcal{L}_{\text{\tiny{APCON-v2}}} = \frac{1}{M} \sum_{i=1}^M \mathcal{L}_{\text{\tiny{APCON-v2}}}^{a^i}.
\end{equation}

\begin{figure}[ht]
    \centering
    \includegraphics[width=0.75\textwidth]{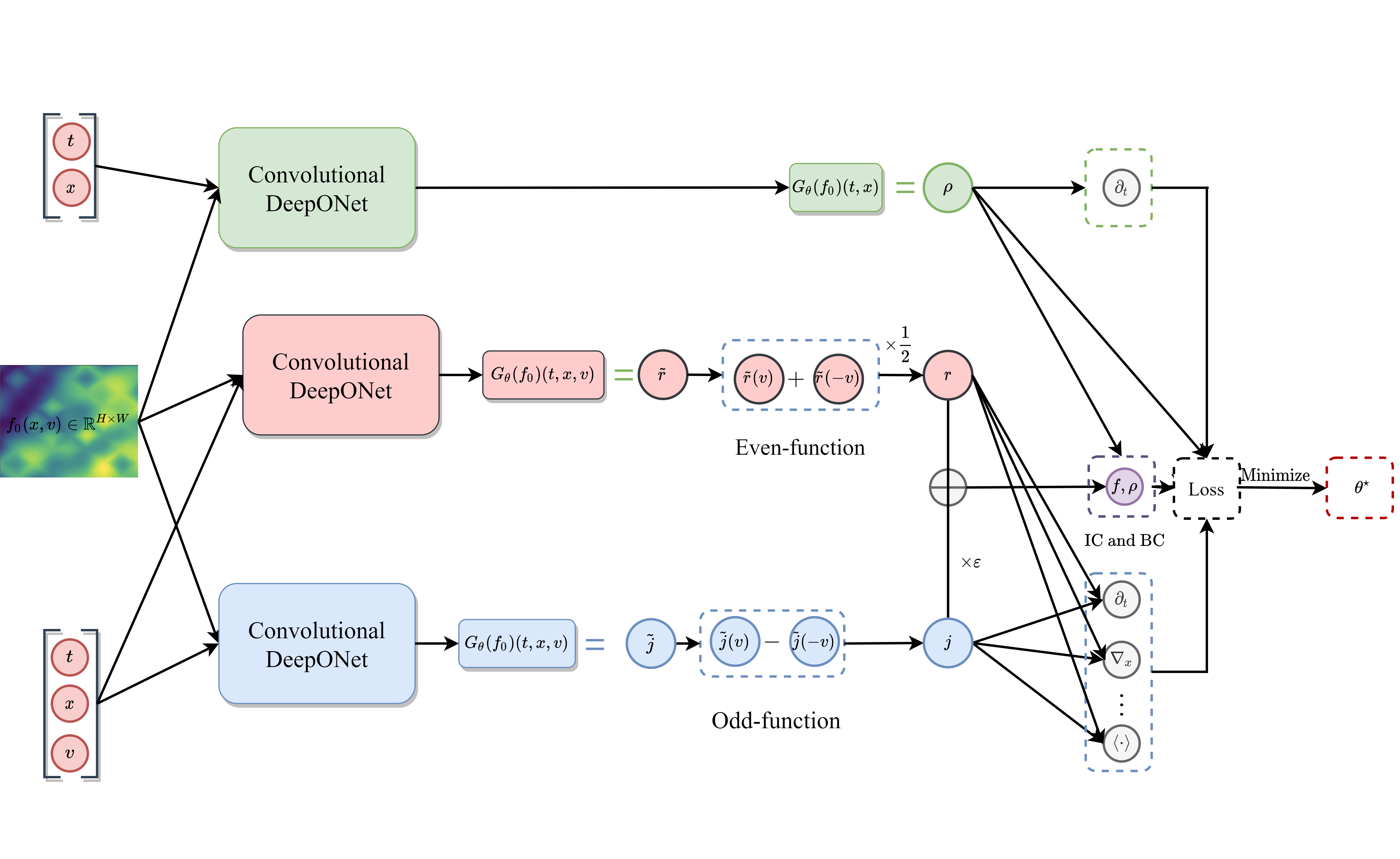}
    \caption{The architecture of Asymptotic-Preserving Convolutional DeepONets based on even-odd decomposition for linear transport equation.}
    \label{fig: apcons-even-odd}
\end{figure}

We put a schematic plot of our APCON method based on even-odd decomposition for the linear transport equation in Figure~\ref{fig: apcons-even-odd}.

\section{Numerical experiments}

The linear transport equations under consideration are accompanied by either inflow or Dirichlet boundary conditions.

Initially, we model random input functions $f_0(x, v)$ by employing a mean-zero Gaussian random field (GRF) with an exponential quadratic covariance kernel:
\begin{equation}\label{eqn: kernel}
    k(\bm{z}, \bm{y}) = \exp \left ( - \frac{\left \| \bm{z} - \bm{y} \right \|^2} {2 l^2} \right ),\, l> 0,
\end{equation}
and, subsequently, we enforce the boundary conditions of the random input functions. In this process, we sample $M = 2^{10}$ distinct discretized initial functions with a shape of $(H, \; W) = (32, \; 64)$ using the GRF method with a covariance kernel parameter of $l = 1/2$. These sampled functions are then divided into training and testing datasets using a split rate of $7:1$.
Furthermore, we approximate the loss using the Monte Carlo method. Specifically, the integration terms involving $\left \langle \cdot \right \rangle$ are computed using the Gauss-Legendre quadrature rule with a total of 32 nodes. During each iteration of an epoch, the batch size for the input function $a$ from the training dataset is set to $B = 4$, resulting in a shape of $(B, H, W)$. Additionally, the number of sample points in the interior domain is $N_{\text{int}} = 2^{10}$, while the number of sample points on the boundary is $N_{\text{bdy}} = 2^{8}$. As for the initial loss, we select the coordinates of the discretized initial functions as sample points, with a total size of $N_0$.
Therefore, for instance, when the inputs consist of a batch of discretized initial functions with a shape of $(B, H, W)$, and the collocations of coordinates have a shape of $(N_{\text{int}}, 3)$, the outputs correspond to the approximate solutions at these coordinates for the $B$ discretized initial functions, resulting in a shape of $(B, N_{\text{int}})$.
The neural networks are trained for a total of $5000$ epochs by minimizing the loss using the Adam optimizer~\cite{kingma2014adam}. The initial learning rate is set to $10^{-4}$, and an annealing schedule is applied with a decay rate of $0.96$ for every $100$ epochs.
For the modified MLP with layer normalization, the branch nets consist of $L = 5$ hidden layers, while the trunk nets consist of $L = 4$ hidden layers. Each hidden layer contains $64$ units. The activation function used for all hidden layers in the modified MLP without layer normalization is $\text{swish}(x)$.
In most of the numerical experiments, we employ two filter layers, where each layer includes a convolution operation with $4$ channels, a kernel shape of $(2, 2)$, and a stride shape of $(2, 2)$. Additionally, an average pooling operation with a shape of $(2, 2)$ and the same shape stride is applied, followed by an activation function with $\text{gelu}(x)$.
To succinctly convey the rationale, we refer to DeepONet as DON and Convolutional DeepONet as CON.

The empirical risk associated with PICON and APCON methods, which utilize micro-macro and even-odd decompositions for the linear transport equation, is
\begin{equation}\label{eqn: risk-pidon}
    \begin{aligned}
    \mathcal{R}_{\text{\tiny{PICON}}} & = \frac{1}{M} \sum_{i=1}^M \left [ \right. \frac{1}{N_{\text{int}}} \sum_{j=1}^{N_{\text{int}}} |\left ( \eps^2 \partial_t f_\theta + \eps v \cdot \nabla_x f_\theta - \LO f_\theta \right ) [a^i](t^j, x^j, v^j)|^2  \\
                               & \qquad \qquad + \frac{1}{N_{\text{bdy}}} \sum_{j=1}^{N_{\text{bdy}}} | \left ( \mathcal{B} f_\theta \right ) [a^i](t^j, x^j, v^j) |^2  \\
                               & \qquad \qquad + \frac{1}{N_{\text{0}}} \sum_{j=1}^{N_{\text{0}}} |\left ( f_\theta - f_0 \right )[a^i](0, x^j, v^j)|^2 \left. \right ],          
    \end{aligned}
\end{equation}
{
\begin{equation}\label{eqn: risk-apcon-v1}
    \begin{aligned}
    \mathcal{R}_{\text{\tiny{APCON-v1}}} & = \frac{1}{M} \sum_{i=1}^M \left [ \right. \frac{1}{N_{\text{int}}} \sum_{j=1}^{N_{\text{int}}} |\left (\partial_t \rho_\theta + \left \langle v \cdot \nabla_x  g_\theta \right \rangle \right ) [a^i](t^j, x^j)|^2  \\
                               & \qquad \qquad + \frac{1}{N_{\text{int}}} \sum_{j=1}^{N_{\text{int}}} |\left ( \eps^2 \partial_t g_\theta + \eps (I - \Pi)({v} \cdot \nabla_{{x}} g_\theta) + {v} \cdot \nabla_{{x}} \rho_\theta - \mathcal{L} g_\theta \right ) [a^i](t^j, x^j, v^j)|^2 \\
                               & \qquad \qquad + \frac{1}{N_{\text{bdy}}} \sum_{j=1}^{N_{\text{bdy}}} | \left ( \mathcal{B} (\rho_\theta + \eps g_\theta) \right ) [a^i](t^j, x^j, v^j) |^2  \\
                               & \qquad \qquad + \frac{1}{N_{\text{0}}} \sum_{j=1}^{N_{\text{0}}} |\left ( (\rho_\theta + \eps g_\theta) - f_0 \right )[a^i](0, x^j, v^j)|^2 \left. \right ],                 
    \end{aligned}
\end{equation}
}
and
\begin{equation}\label{eqn: risk-apcon-v2}
    \begin{aligned}
    \mathcal{R}_{\text{\tiny{APCON-v2}}} & = \frac{1}{M} \sum_{i=1}^M \left [ \right. \frac{1}{N_{\text{int}}} \sum_{j=1}^{N_{\text{int}}} |\left ( \eps^2 \partial_t r_\theta + \eps^2 v \cdot \nabla_x  j_\theta - (\rho_\theta - r_\theta) \right ) [a^i](t^j, x^j, v^j)|^2  \\
                               & \qquad \qquad + \frac{1}{N_{\text{int}}} \sum_{j=1}^{N_{\text{int}}} |\left ( \eps^2 \partial_t j_\theta + v \cdot \nabla_x  r_\theta - ( - j_\theta) \right ) [a^i](t^j, x^j, v^j)|^2 \\
                               & \qquad \qquad + \frac{1}{N_{\text{int}}} \sum_{j=1}^{N_{\text{int}}} |\left ( \partial_t \rho_\theta + \left \langle v \cdot \nabla_x j_\theta \right \rangle \right ) [a^i](t^j, x^j)|^2 \\
                               & \qquad \qquad + \frac{1}{N_{\text{int}}} \sum_{j=1}^{N_{\text{int}}} |\left ( \rho_\theta - \left \langle j_\theta \right \rangle \right ) [a^i](t^j, x^j)|^2 \\
                               & \qquad \qquad + \frac{1}{N_{\text{bdy}}} \sum_{j=1}^{N_{\text{bdy}}} | \left ( \mathcal{B} (r_\theta + \eps j_\theta) \right ) [a^i](t^j, x^j, v^j) |^2  \\
                               & \qquad \qquad + \frac{1}{N_{\text{0}}} \sum_{j=1}^{N_{\text{0}}} |\left ( (r_\theta + \eps j_\theta) - f_0 \right )[a^i](0, x^j, v^j)|^2 \left. \right ],                 
    \end{aligned}
\end{equation}
respectively.

The reference solutions are obtained by fast spectral method and
we will check the relative $\ell^2$ error of the density solution $\rho(t, x)$ at all uniform grids $\{(t_i, x_j)\}_{1 \le i \le I, 1 \le j \le H}$,
\begin{equation}
    \text{error} := \sqrt{
    \frac{\sum_{i,j} |\rho_{\theta}(t_i, x_j) - \rho_{\text{ref}}(t_i, x_j)|^2}{\sum_{i,j} |\rho_{\text{ref}}(t_i, x_j)|^2}
    },
\end{equation}
where $\rho_{\theta}$ is the neural operator solution approximation, and $\rho_{\text{ref}}$ is the reference solution.

Subsequently, we will conduct several experiments to compare the performance of PIDONs/CONs and APDONs/CONs utilizing the micro-macro and even-odd decompositions.
The relative $\ell^2$ error below is the optimal result obtained after $5$ trials in all experiments.

\subsection{Problem I: Inflow condition}

Consider the linear transport equation in a boundary domian $\mathcal{T} \times \mathcal{D} \times {\Omega} = [0, T] \times [0, 1] \times [-1, 1]$:
\begin{equation}\label{eqn: lte-eps-1}
    \eps \partial_t f + v \cdot \nabla_x f = \frac{1}{\eps} \left ( \frac{1}{2} \int_{-1}^1 f \, \diff v - f \right ),
\end{equation}
with in-flow boundary conditions as
\begin{equation}
    \begin{aligned}
        f(t, x_L, v) & = 1, \quad \text{for} \quad v > 0, \\
        f(t, x_R, v) & = \frac{1}{2}, \quad \text{for} \quad v < 0.
    \end{aligned}
\end{equation}
Our focus is on the operator learning task that maps the initial function $f_0(x, v)$ to the solution $f(t, x, v)$ for the entire domain $\mathcal{T} \times \mathcal{D} \times {\Omega}$. 
Initially, to generate initial conditions that comply with the inflow boundary condition, we sample $\tilde{f_0}(x, v) \sim \text{GRF}$ and select positive samples to construct
\begin{equation}\label{eqn: initial-funtions}
    f_0(x, v) = \left ( \text{relu}^3(v) \cdot x + \text{relu}^3(-v) \cdot (1 - x) \right ) \cdot \tilde{f_0}(x, v) + (1 - \frac{1}{2} x).
\end{equation}

\subsubsection{Kinetic regime with $\eps = 1$}

\begin{figure}[htbp!]
	\centering
	\subfigure[PIDON]{
		\includegraphics[width=0.5\textwidth]{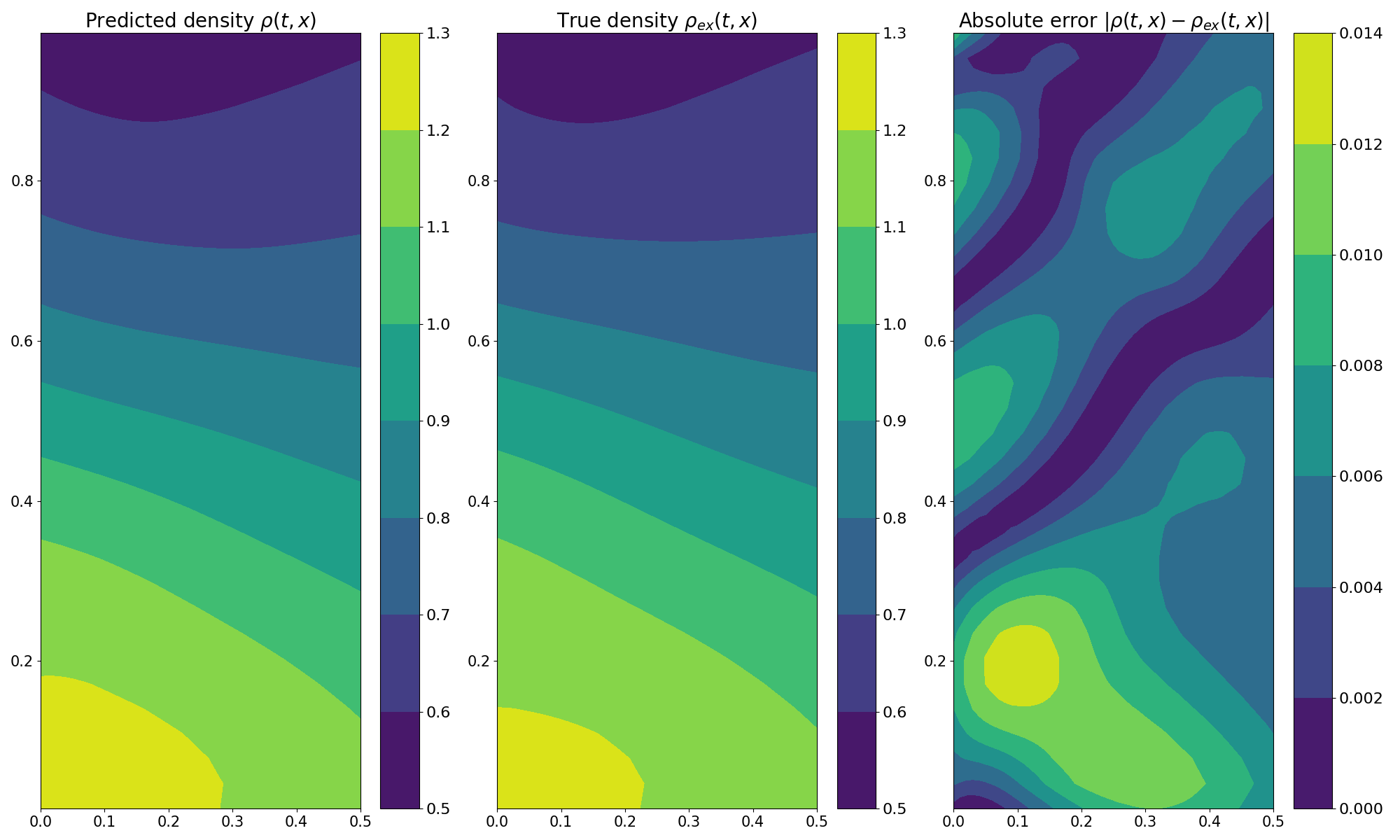}
	}  
	\subfigure[APDON-v1]{
		\includegraphics[width=0.5\textwidth]{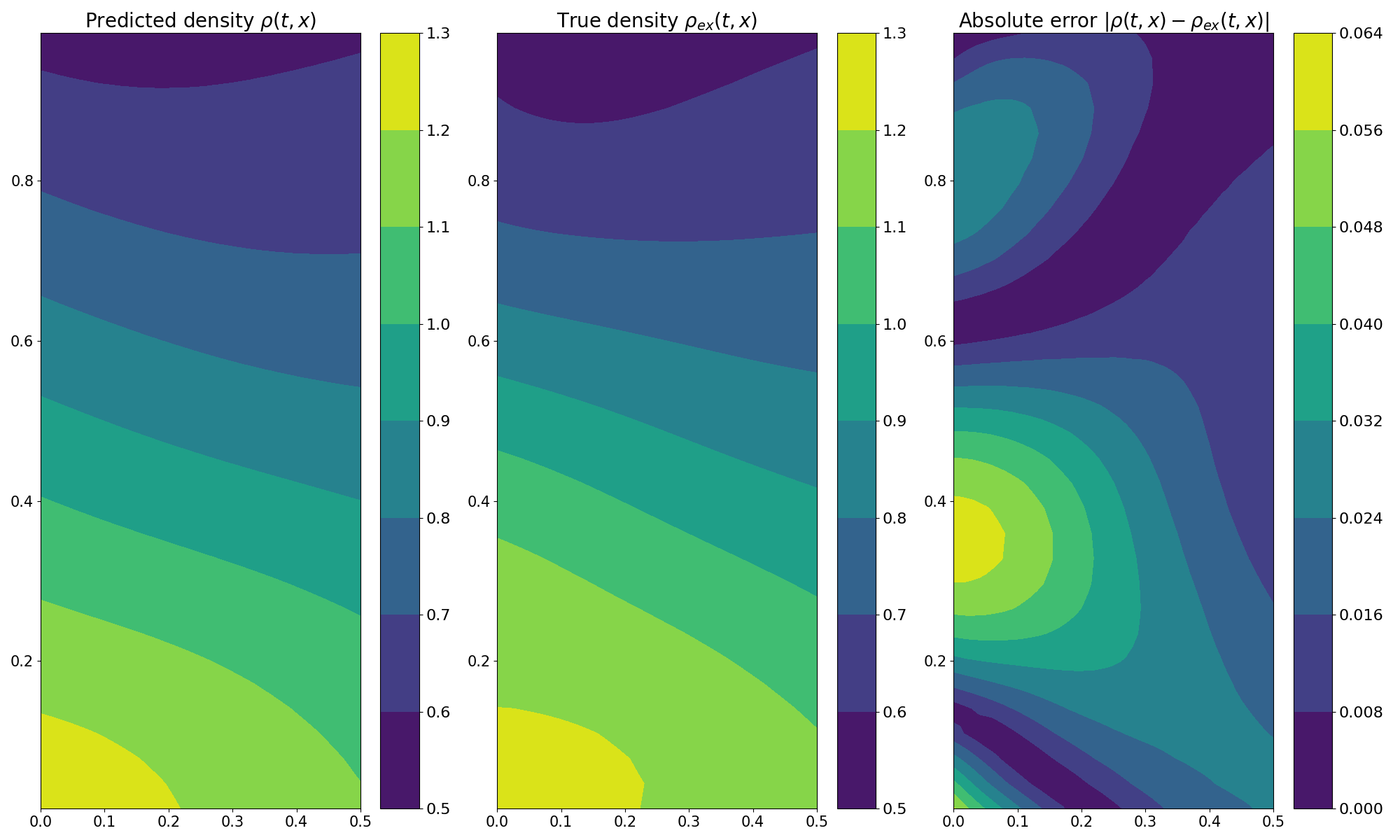}
	}  
	\subfigure[APDON-v2]{
		\includegraphics[width=0.5\textwidth]{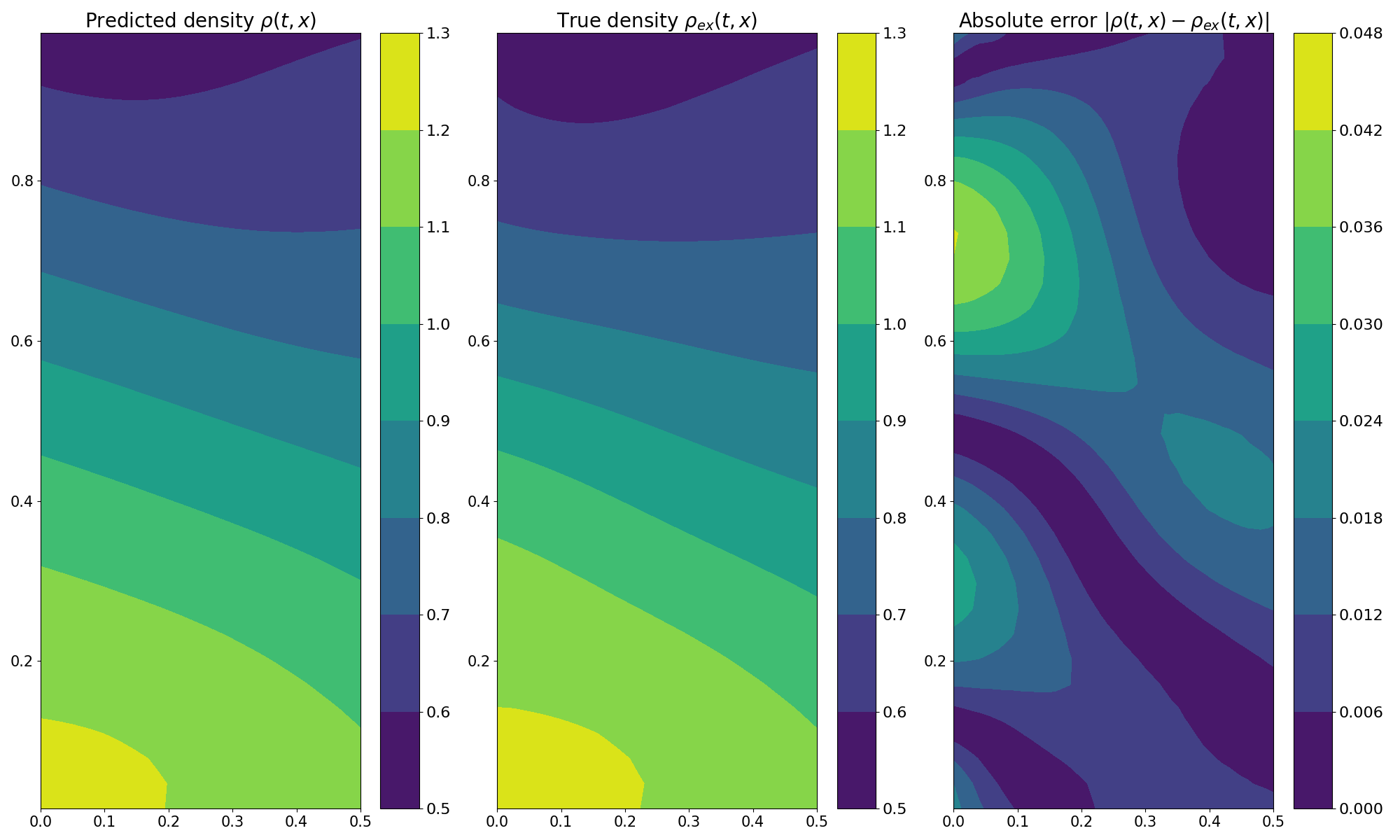}
	} 
\caption{Plots of density $\rho(t, x)$ in $(t, x) \in [0, 0.5] \times [0, 1]$ by PIDON, APDON-v1, APDON-v2 for a representative input function. The results are obtained by DeepONets equipped with modified MLP. }
\label{fig: lte-don-1e0}
\end{figure}

\begin{figure}[htbp!]
	\centering
	\subfigure[PICON]{
		\includegraphics[width=0.5\textwidth]{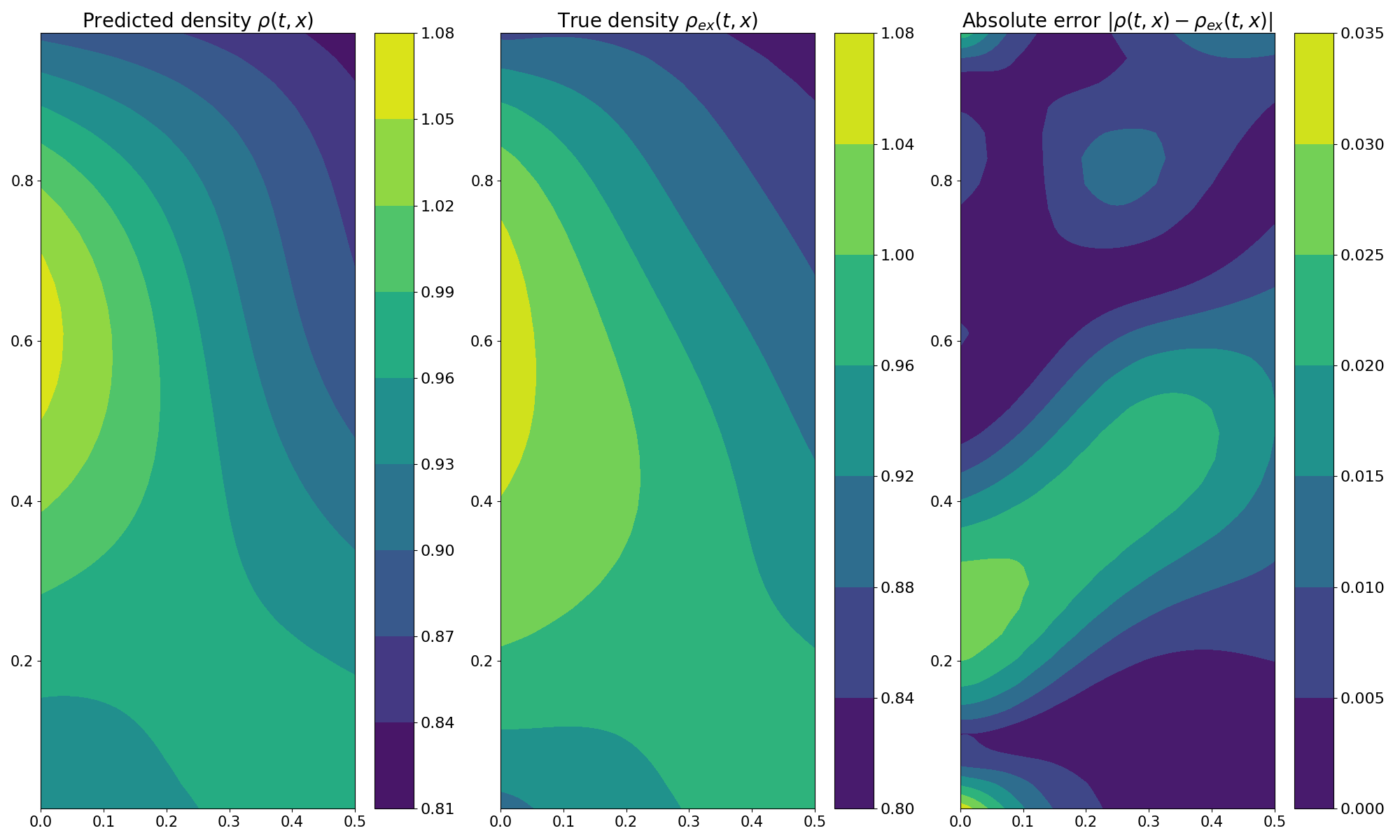}
	}  
	\subfigure[APCON-v1]{
		\includegraphics[width=0.5\textwidth]{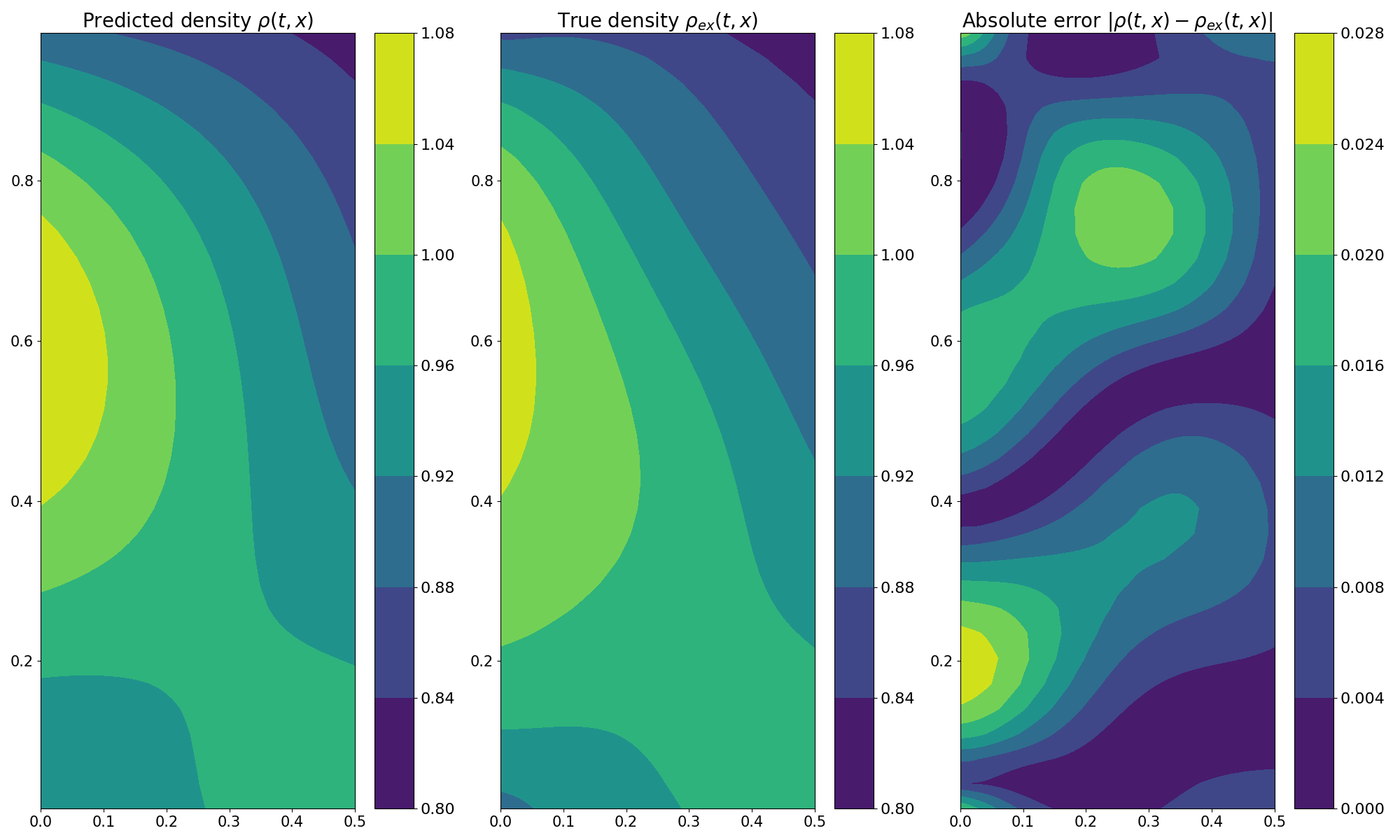}
	}  
	\subfigure[APCON-v2]{
		\includegraphics[width=0.5\textwidth]{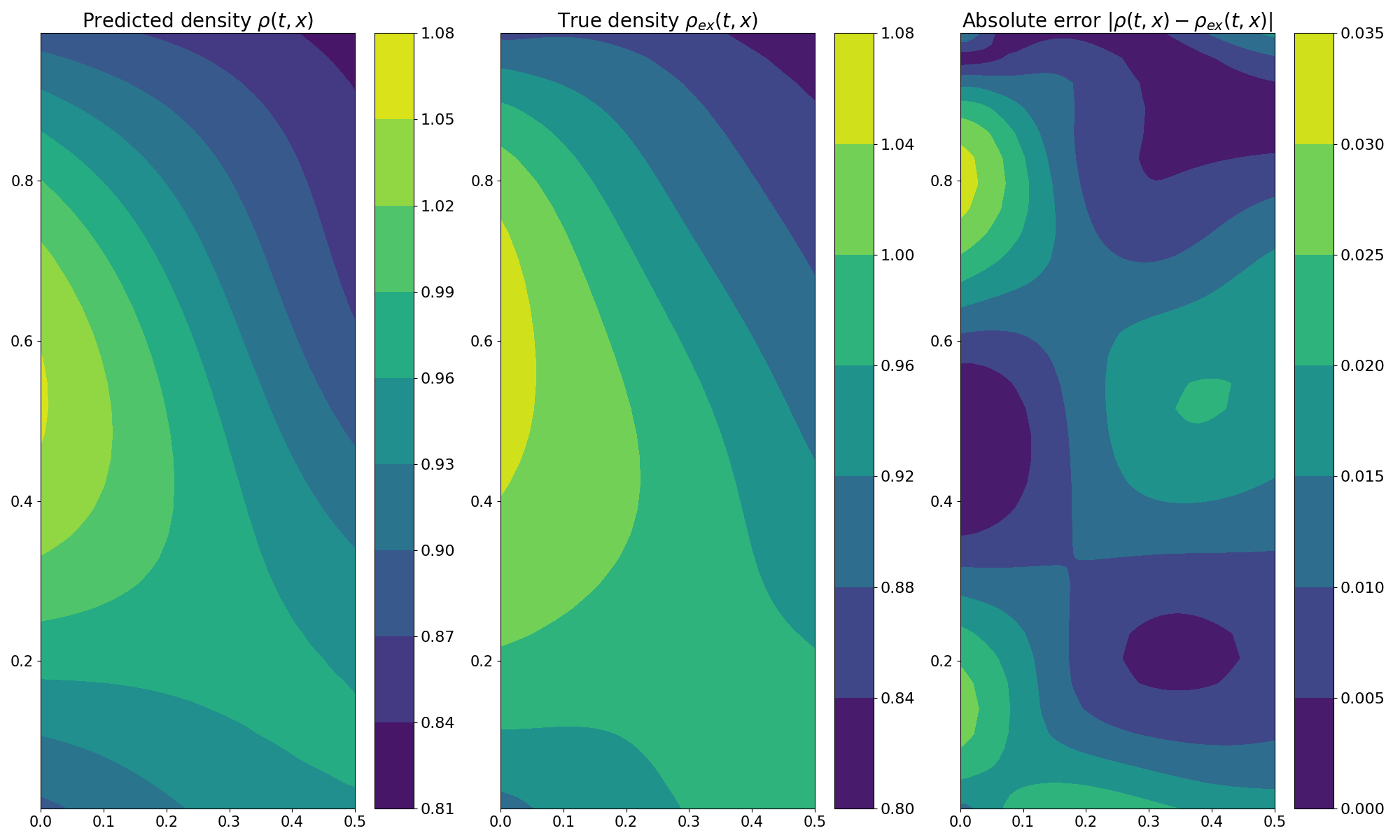}
	} 
\caption{Plots of density $\rho(t, x)$ in $(t, x) \in [0, 0.5] \times [0, 1]$ by PICON, APCON-v1, APCON-v2 method for a representative input function. The results are obtained by convolutional DeepONets equipped with modified MLP and layer normalization. }	
\label{fig: lte-con-1e0}
\end{figure}

We plot the density predictions made by PIDON, APDON-v1, APDON-v2 and PICON, APCON-v1, APCON-v2 for a representative input function from the test datasets, alongside the reference solution in Figures~\ref{fig: lte-don-1e0} and \ref{fig: lte-con-1e0}, respectively.

\begin{figure}[ht]
\centering
\includegraphics[width=0.75\textwidth]{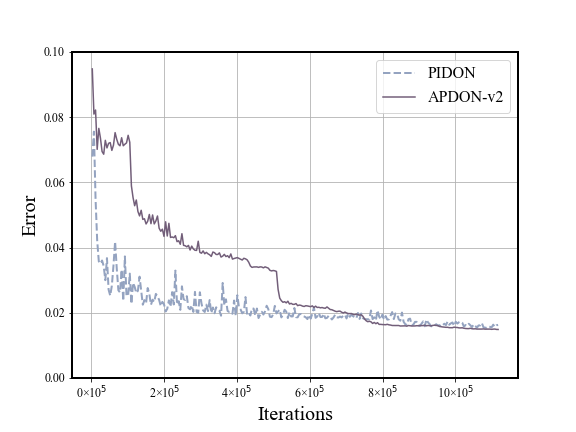}
\caption{The relative $l^2$ error of PIDON and APDON-v2 in the kinetic regime ($\eps = 1$) of Problem I.}
\label{fig: error}
\end{figure}

Table~\ref{tab: errors-1} records the relative $\ell^2$ error of the density solution and the number of parameters for these methods. Our quantities of interest (QoIs) include the relative $\ell^2$ error and the number of parameters for these neural networks. {
The relative $l^2$ error during the training of PIDON and APDON-v2 in the kinetic regime ($\eps = 1$) is depicted in Figure~\ref{fig: error}. It is evident that PIDON converges more swiftly than our method, primarily owing to its reduced parameter count and the lack of diffusion effects in this scenario.}
\begin{small}
    \begin{table}[tbhp]
        \caption{Comparison  in  kinetic regime ($\eps = 1$) of Problem I.}\label{tab: errors-1}
        \centering
        \begin{tabular}{ccccccc}
            \toprule[1pt]
            \noalign{\smallskip}
            \multirow{2}*{\diagbox{QoIs}}{{Method}}
             & \multicolumn{2}{c}{${\text{PI}}$} & \multicolumn{2}{c}{${\text{AP v1}}$} & \multicolumn{2}{c}{${\text{AP v2}}$}                                                                                                                        \\
             &\multicolumn{1}{c}{DON}  & \multicolumn{1}{c}{CON}     & \multicolumn{1}{c}{DON}  & \multicolumn{1}{c}{CON} & \multicolumn{1}{c}{DON} & \multicolumn{1}{c}{CON}  \\
            \noalign{\smallskip}
            \midrule[1pt]
            \noalign{\smallskip}
            \multirow{1}*{{{Relative $\ell^2$ error}}}
             &  ${\text{1.88 e-}2}$ & ${\text{1.74 e-}2}$ & ${\text{3.32 e-}2}$ & ${\text{1.59 e-}2}$ & ${\text{1.48 e-}2}$  & ${\text{1.44 e-}2}$     \\     
            \multirow{1}*{{{$\#$ params}}}
             & ${\text{423 296}}$   & ${\text{43 288}}$  &  ${\text{846 400}}$  &  ${\text{86 384}}$   &  ${\text{1 269 696}}$  & ${\text{129 672}}$     \\     
             
            \noalign{\smallskip}
            \bottomrule[1pt]
        \end{tabular}
    \end{table}
\end{small}

\subsubsection{Diffusion regime with $\eps = 10^{-4}$}

Table~\ref{tab: errors-2} records the relative $\ell^2$ error of the density solution for these methods. The number of parameters for these methods is the same as previously mentioned. Here, a ``-'' indicates that the training process does not converge after reaching the maximum number of training epochs. This suggests that the loss function of operator learning in the form of vanilla PINN disregards the asymptotic property of the problem's multiscale nature and fails to capture the macroscopic behavior corresponding to small physical parameters, as explained in \cite{wuAPNN}.

\begin{small}
    \begin{table}[tbhp]
        \caption{Comparison in  the diffusion regime ($\eps = 10^{-4}$) of Problem I.}\label{tab: errors-2}
        \centering
        \begin{tabular}{ccccccc}
            \toprule[1pt]
            \noalign{\smallskip}
            \multirow{2}*{\diagbox{QoI}}{{Method}}
             & \multicolumn{2}{c}{${\text{PI}}$} & \multicolumn{2}{c}{${\text{AP v1}}$} & \multicolumn{2}{c}{${\text{AP v2}}$}                                                                                                                        \\
             &\multicolumn{1}{c}{DON}  & \multicolumn{1}{c}{CON}     & \multicolumn{1}{c}{DON}  & \multicolumn{1}{c}{CON} & \multicolumn{1}{c}{DON} & \multicolumn{1}{c}{CON}  \\
            \noalign{\smallskip}
            \midrule[1pt]
            \noalign{\smallskip}
            \multirow{1}*{{{Relative $\ell^2$ error}}}
             &  $-$ & $-$ & ${\text{ 2.63e-}2}$ & ${\text{ 1.58e-}2}$ & ${\text{ 4.09e-}2}$  & ${\text{1.55 e-}2}$     \\                     
            \noalign{\smallskip}
            \bottomrule[1pt]
        \end{tabular}
    \end{table}
\end{small}

\begin{figure}[htbp!]
	\centering
	\subfigure[APCON-v1]{
		\includegraphics[width=0.6\textwidth]{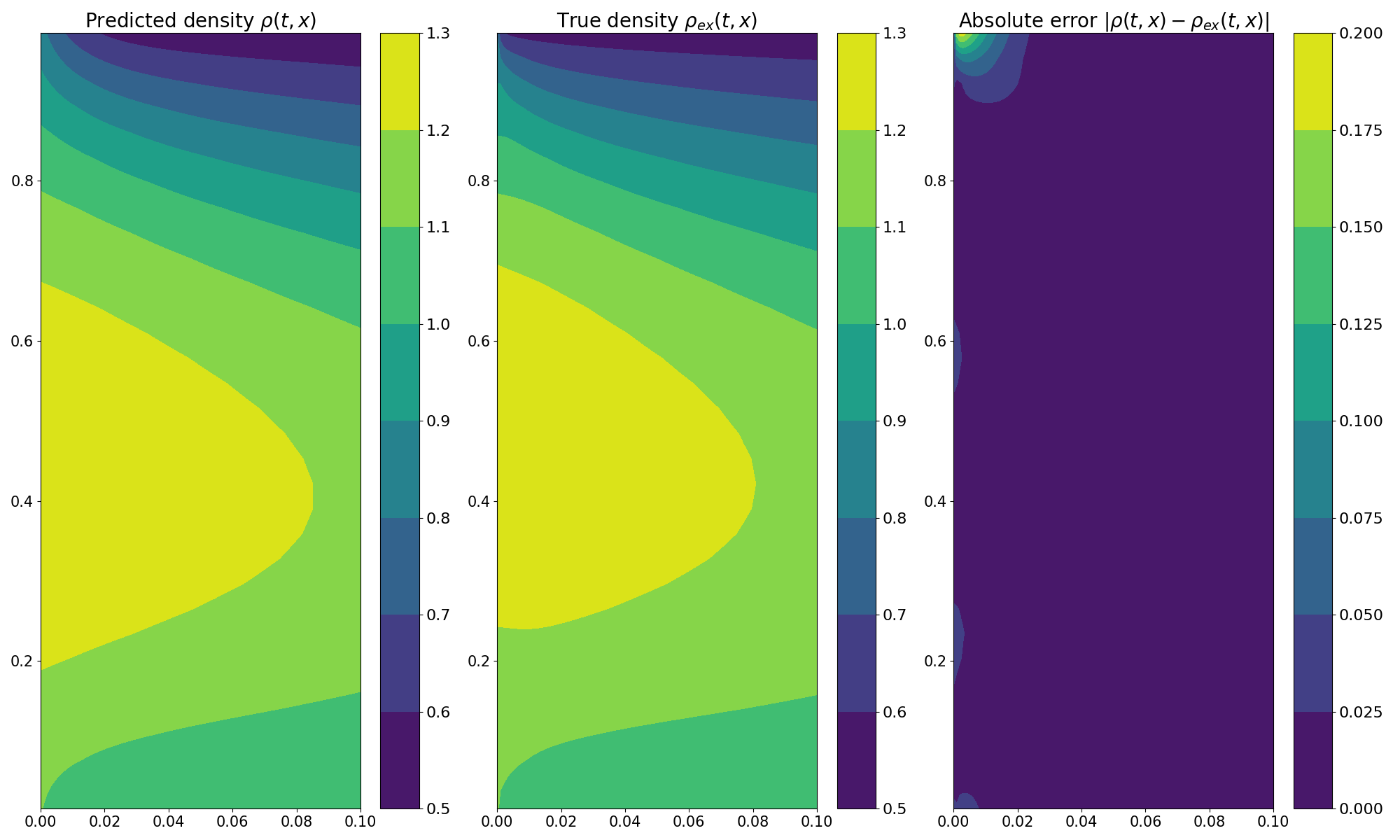}
	}  
	\subfigure[APCON-v2]{
		\includegraphics[width=0.6\textwidth]{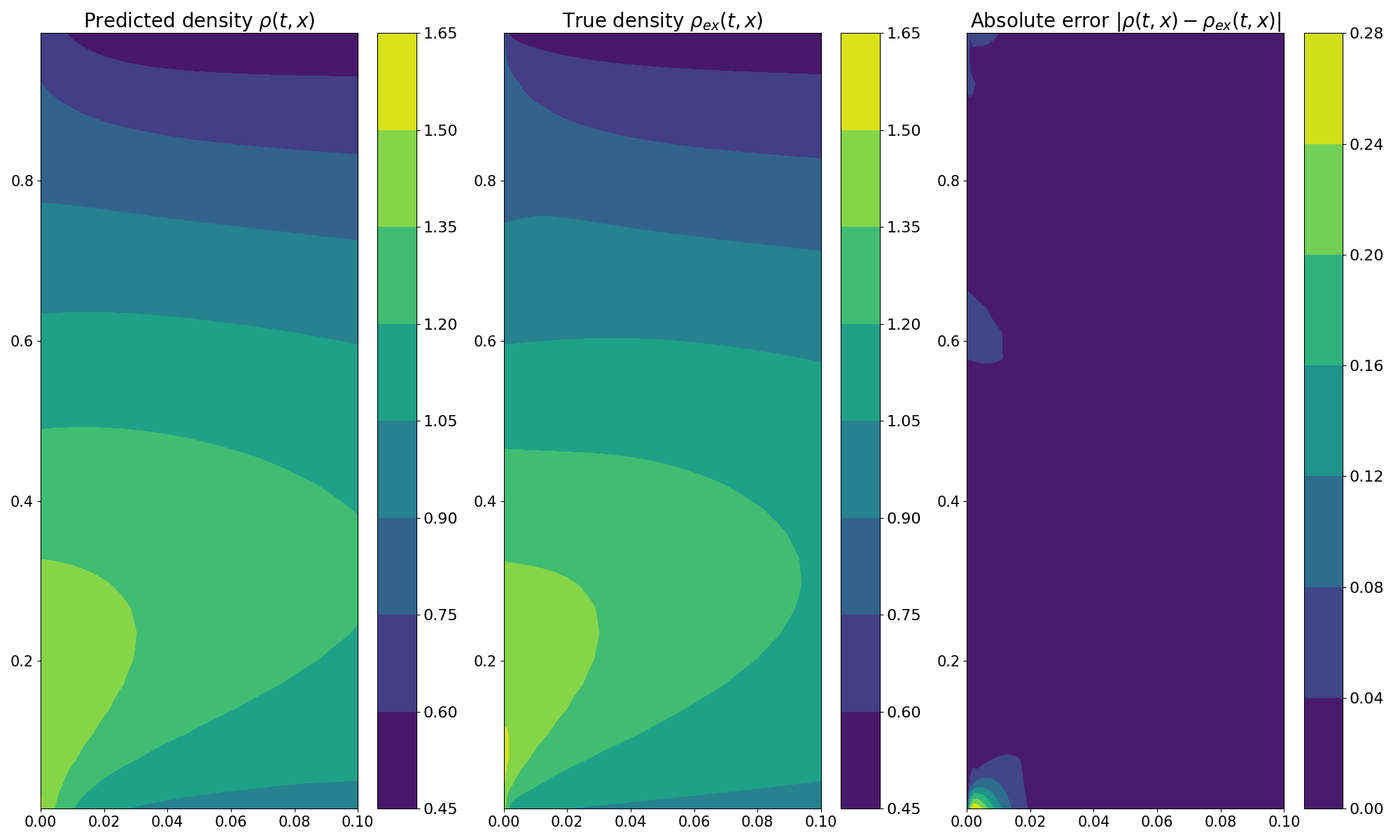}
	} 
\caption{Plots of density $\rho(t, x)$ in $(t, x) \in [0, 0.1] \times [0, 1]$ by APCON-v1 and APCON-v2 method for two representative input functions. The results are obtained by convolutional DeepONets equipped with modified MLP and layer normalization. }	
\label{fig: lte-con-1e-4}
\end{figure}

We plot the density predictions made by APCON-v1 and APCON-v2 for two representative input functions from the test datasets, alongside the reference solution in Figure~\ref{fig: lte-con-1e-4}.

\subsection{Problem II: Dirichlet condition}

Consider the linear transport equation in a boundary {domain} $[0, 0.1] \times [0, 1] \times [-1, 1]$ with $\eps = 10^{-4}$:
\begin{equation}\label{eqn: lte-eps-2}
    \eps \partial_t f + v \cdot \nabla_x f = \frac{1}{\eps} \left ( \frac{1}{2} \int_{-1}^1 f \, \diff v - f \right ),
\end{equation}
with the  boundary conditions $f(t, x_L, v) = 0 = f(t, x_R, v)$.
The initial function is chosen from a distinct function space with a Maxwellian form as
\begin{equation}\label{eqn: sample-f0}
    f_0 = r \cdot \left [ 1 + \sin(2 \pi x - \frac{\pi}{2}) \right ] \cdot 3 \exp \left ( - \frac{(3v)^2}{2}\right ), 
\end{equation}
where $r$ is sampled from $[0, 1]$.

\begin{figure}[htbp!]
	\centering
	\subfigure[APCON-v1]{
		\includegraphics[width=0.6\textwidth]{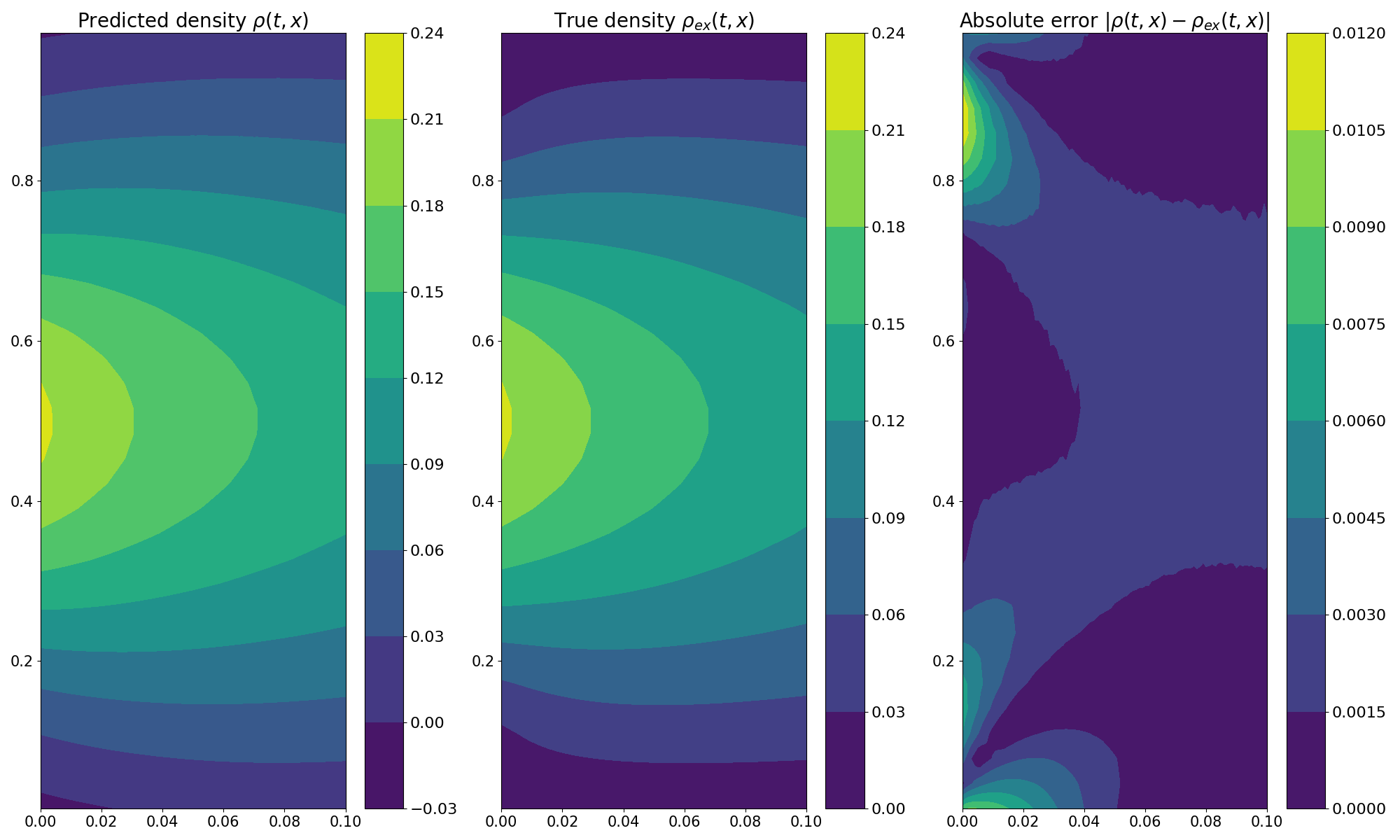}
	}  
	\subfigure[APCON-v2]{
		\includegraphics[width=0.6\textwidth]{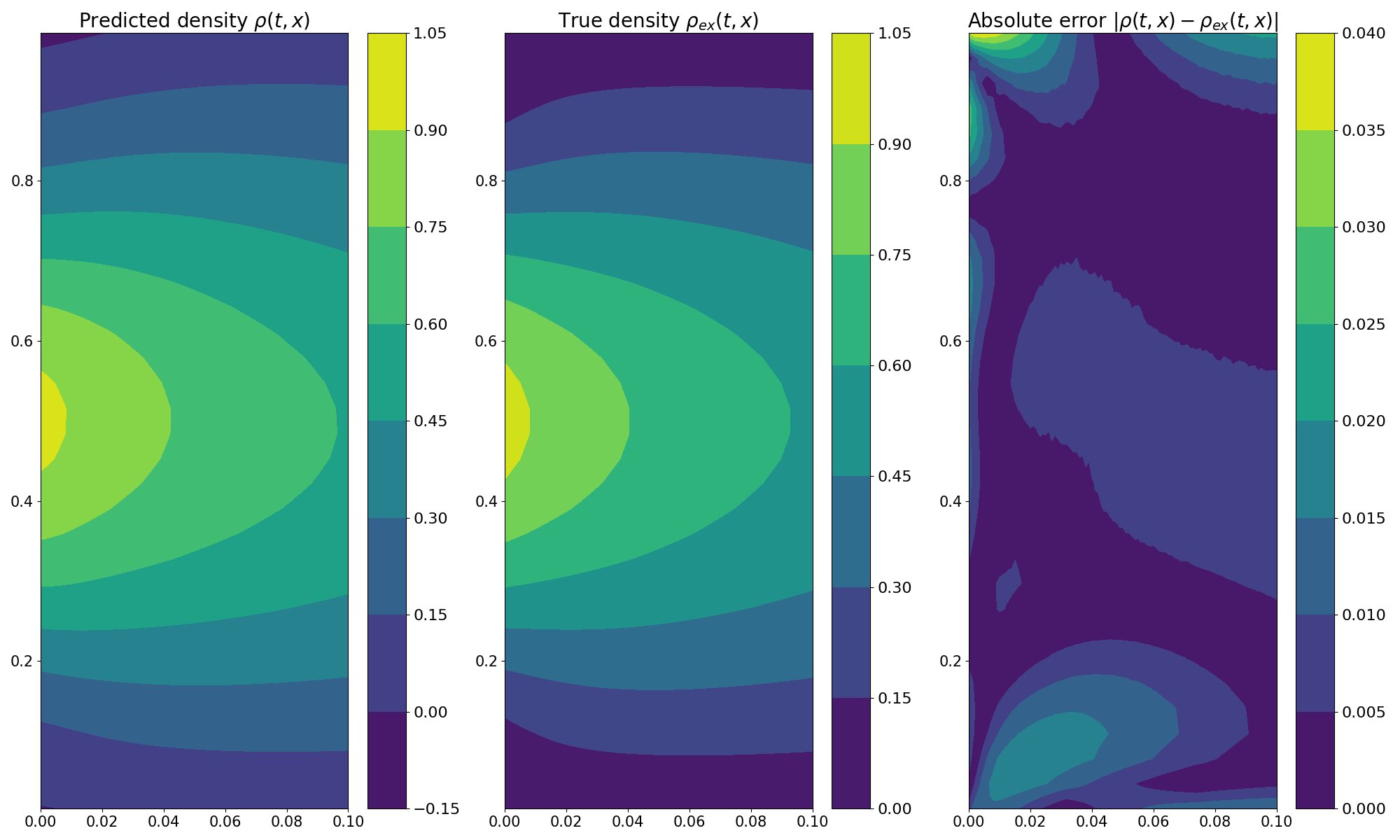}
	} 
\caption{Plots of density $\rho(t, x)$ in $(t, x) \in [0, 0.1] \times [0, 1]$ by APCON-v1 and APCON-v2 method for two representative input functions. The results are obtained by convolutional DeepONets equipped with modified MLP and layer normalization. }	
\label{fig: lte-con-1e-4-maxwell}
\end{figure}

We plot the density predictions made by APCON-v1 and APCON-v2 for two representative input functions from the test datasets, alongside the reference solution in Figure~\ref{fig: lte-con-1e-4-maxwell}.

Table~\ref{tab: errors-3} records the relative $\ell^2$ error of the density solution for these methods.

\begin{small}
    \begin{table}[tbhp]
        \caption{Comparison in  the diffusion regime ($\eps = 10^{-4}$) of Problem II.}\label{tab: errors-3}
        \centering
        \begin{tabular}{ccc}
            \toprule[1pt]
            \noalign{\smallskip}
            \multirow{2}*{\diagbox{QoI}}{{Method}}
             & \multicolumn{1}{c}{${\text{APCON-v1}}$} & \multicolumn{1}{c}{${\text{APCON-v2}}$}     \\
             &   &    \\
            \noalign{\smallskip}
            \midrule[1pt]
            \noalign{\smallskip}
            \multirow{1}*{{{Relative $\ell^2$ error}}}
             & ${\text{1.67 e-}2}$ & ${\text{1.62 e-}2}$     \\                     
            \noalign{\smallskip}
            \bottomrule[1pt]
        \end{tabular}
    \end{table}
\end{small}

\subsection{Computational cost}

\begin{figure}[ht]
    \centering
    \includegraphics[width=0.75\textwidth]{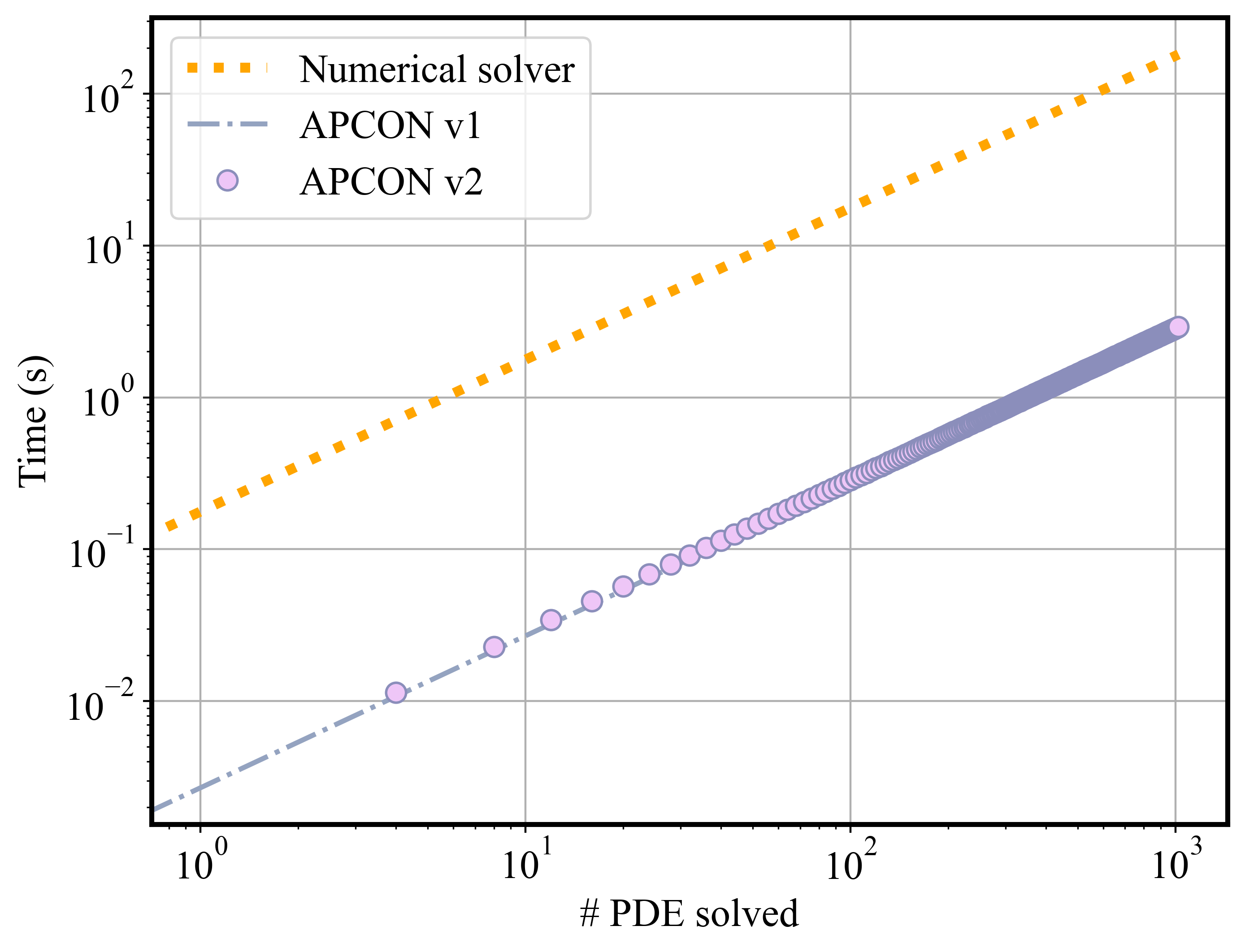}
    \caption{Computational cost (second) for performing inference with pre-trained APCONs, along with the corresponding time required to solve a PDE with a conventional fast spectral method which encompasses a scalable GPU implementation utilizing cupy.
    %It is worth highlighting that our APCON models are capable of predicting the solution for $2^{10}$ time-dependent PDEs within a mere two milliseconds, achieving a performance improvement of up to two orders of magnitude when compared to a conventional PDE solver.
    Reported timings are obtained on a single NVIDIA RTX 3090 graphics processing unit (GPU). The average total time costs for the classical numerical solver, APCON-v1, and APCON-v2 are $176.9$ ms, $2.69$ ms, and $2.85$ ms, respectively.}
    \label{fig: computational-cost}
\end{figure}

In this section, we present the comparative analysis of computational costs between APCONs and a numerical solver for the diffusion regime ($\eps = 10^{-4}$) in Problem I. Our APCONs are implemented using JAX with the Haiku library, while the numerical solver is a fast spectral method with a scalable implementation on a graphics processing unit (GPU) using cupy. As depicted in Figure~\ref{fig: computational-cost}, {our APCON methods demonstrate exceptional performance by accurately predicting the solutions and the average time is about 2 ms. It is worth noting that, even though the APCON methods offer an advantage in terms of inference speed, the training process for APCON methods tends to be relatively time-consuming. In this case, the training time for APCON-v1 and APCON-v2 amounts to 49907 and 46783 seconds, respectively.} This  represents a remarkable advancement, surpassing the capabilities of conventional PDE solvers by up to two orders of magnitude. Moreover, the inference process with APCONs is effortlessly parallelizable, enabling the prediction of solutions for an even larger number of equations using multiple GPUs.

\subsection{Layer normalization}

In this section, we investigate the relative $\ell^2$ error without layer normalization for the diffusion regime ($\eps = 10^{-4}$) of Problem I.

Table~\ref{tab: errors-4} documents the relative $\ell^2$ error of the density solution.
It is worth noting that the utilization of layer normalization has the potential to enhance the precision of APCON-v2.
\begin{small}
    \begin{table}[tbhp]
        \caption{APCON-v2 w/o LayerNorm for the diffusion regime ($\eps = 10^{-4}$) of Problem I.}\label{tab: errors-4}
        \centering
        \begin{tabular}{ccc}
            \toprule[1pt]
            \noalign{\smallskip}
            \multirow{2}*{\diagbox{QoI}}{{Method}}
             & \multicolumn{1}{c}{${\text{without LayerNorm}}$}   & \multicolumn{1}{c}{${\text{with LayerNorm}}$}  \\
             &   &  \\
            \noalign{\smallskip}
            \midrule[1pt]
            \noalign{\smallskip}
            \multirow{1}*{{{Relative $\ell^2$ error}}}
             & ${\text{2.38 e-}2 }$  & ${\text{1.55 e-}2}$     \\                     
            \noalign{\smallskip}
            \bottomrule[1pt]
        \end{tabular}
    \end{table}
\end{small}

\subsection{Order of pooling and activation}

Within the filter layers, it is possible to perform the activation step prior to the pooling operation. In this specific scenario, we are examining the sequencing of pooling and activation operations for the kinetic regime ($\eps = 1$) of Problem I, employing APCON-v1.

Table~\ref{tab: errors-order} records the relative $\ell^2$ error of the density solution pertaining to the ordering of pooling and activation, as observed through the utilization of APCON-v1. It is evident from the table that the impact on final accuracy in this particular case is negligible.

\begin{small}
    \begin{table}[tbhp]
        \caption{The order of pooling and activation for the kinetic regime ($\eps = 1$) of Problem I.}\label{tab: errors-order}
        \centering
        \begin{tabular}{ccc}
            \toprule[1pt]
            \noalign{\smallskip}
            \multirow{2}*{\diagbox{QoI}}{{Order}}
             & \multicolumn{1}{c}{${\text{Activation}} + {\text{Pooling}}$} & \multicolumn{1}{c}{${\text{Pooling}} + {\text{Activation}}$}     \\
             &    &    \\
            \noalign{\smallskip}
            \midrule[1pt]
            \noalign{\smallskip}
            \multirow{1}*{{{Relative $\ell^2$ error}}}
             & ${\text{1.66 e-}2}$ & ${\text{1.59 e-}2}$     \\                     
            \noalign{\smallskip}
            \bottomrule[1pt]
        \end{tabular}
    \end{table}
\end{small}

\subsection{Kernel shape and channels}

We shall now proceed to investigate the impact of kernel size and channels in the convolutional layer of APCON-v2, specifically concerning the diffusion regime ($\eps = 10^{-4}$) of Problem I.

Table~\ref{tab: errors-kernel} and Table~\ref{tab: errors-channel} document the relative $\ell^2$ error of the density solution for different kernel shapes, namely $(1, 2)$, $(2, 2)$ and $(2, 4)$, as well as varying numbers of channels, namely $2$, $4$, and $6$. All other settings remain unchanged. It is evident that the kernel size of $(2, 2)$ with $4$ channels demonstrates superior performance.

\begin{small}
    \begin{table}[tbhp]
        \caption{The kernel size of convolutional layer for the kinetic regime ($\eps = 10^{-4}$) of Problem I.}\label{tab: errors-kernel}
        \centering
        \begin{tabular}{cccc}
            \toprule[1pt]
            \noalign{\smallskip}
            \multirow{2}*{\diagbox{QoI}}{{Kernel size}}
             & \multicolumn{1}{c}{$(1, 2)$} & \multicolumn{1}{c}{$(2, 2)$}  & \multicolumn{1}{c}{$(2, 4)$}   \\
             &    &    & \\
            \noalign{\smallskip}
            \midrule[1pt]
            \noalign{\smallskip}
            \multirow{1}*{{{Relative $\ell^2$ error}}}
             & ${\text{2.52 e-}2}$ & ${\text{1.55 e-}2}$ &  ${\text{2.51 e-}2}$ \\                     
            \noalign{\smallskip}
            \bottomrule[1pt]
        \end{tabular}
    \end{table}
\end{small}

\begin{small}
    \begin{table}[tbhp]
        \caption{The number of channels for the kinetic regime ($\eps = 10^{-4}$) of Problem I.}\label{tab: errors-channel}
        \centering
        \begin{tabular}{cccc}
            \toprule[1pt]
            \noalign{\smallskip}
            \multirow{2}*{\diagbox{QoI}}{$\#$ {Channels}}
             & \multicolumn{1}{c}{$2$} & \multicolumn{1}{c}{$4$}  & \multicolumn{1}{c}{$6$}   \\
             &    &    & \\
            \noalign{\smallskip}
            \midrule[1pt]
            \noalign{\smallskip}
            \multirow{1}*{{{Relative $\ell^2$ error}}}
             & ${\text{2.78 e-}2}$ & ${\text{1.55 e-}2}$ &  ${\text{2.18 e-}2}$ \\                     
            \noalign{\smallskip}
            \bottomrule[1pt]
        \end{tabular}
    \end{table}
\end{small}

\subsection{Filter layers}

Lastly, we investigate the impact of the number of filter layers on the diffusion regime ($\eps = 10^{-4}$) of Problem I. Table~\ref{tab: errors-filters} illustrates the relative $\ell^2$ error of the density solution when utilizing a single and two filter layers compared to other identical settings. 

\begin{small}
    \begin{table}[tbhp]
        \caption{The number of filter layers for the kinetic regime ($\eps = 10^{-4}$) of Problem I.}\label{tab: errors-filters}
        \centering
        \begin{tabular}{cccc}
            \toprule[1pt]
            \noalign{\smallskip}
            \multirow{2}*{\diagbox{QoI}}{$\#$ {Filter layers}}
             & \multicolumn{1}{c}{$1$} & \multicolumn{1}{c}{$2$}   \\
             &    &    \\
            \noalign{\smallskip}
            \midrule[1pt]
            \noalign{\smallskip}
            \multirow{1}*{{{Relative $\ell^2$ error}}}
             & ${\text{2.30 e-}2}$ & ${\text{1.55 e-}2}$  \\                     
            \noalign{\smallskip}
            \bottomrule[1pt]
        \end{tabular}
    \end{table}
\end{small}

\section{Conclusion}

In this paper, we focus on learning the numerical solution operator of multiscale linear transport equations that involve diffusive scaling.
We observed that the vanilla physics-informed DeepONets, when equipped with a modified MLP, may encounter instability issues in maintaining the desired macroscopic behavior. 
To tackle this challenge, we first introduce the concept of an Asymptotic-Preserving (AP) loss function within the realm of operator learning. 
Subsequently, drawing inspiration from the heat kernel in the diffusion equation, we propose an innovative architecture called Convolutional DeepONets. 
These networks leverage multiple local convolution operations, rather than relying on a global heat kernel, while incorporating pooling and activation operations within each filter layer. 
As a result, we present two novel variants of Asymptotic-Preserving Convolutional DeepONets (APCONs) based on micro-macro and even-odd decomposition. 
Significantly, our APCON approaches maintain a consistent parameter count regardless of the grid size, effectively capturing the inherent diffusive behavior in the linear transport problem.

\section*{Acknowledgement}
SJ was partially supported by the National Key R\&D Program of China (no.
2020YFA0712000), the NSFC grant No. 12031013, and the Shanghai Municipal Science
and Technology Major Project (2021SHZDZX0102).
Shi Jin is also supported by NSFC grant No. 11871297. Zheng Ma is also supported by NSFC Grant No. 12031013, No.92270120 and partially supported by Institute of Modern Analysis---A Shanghai Frontier Research Center.

\bibliographystyle{plain}
\bibliography{ref.bib}

\end{document}